\documentclass[journal]{IEEEtran}
\ifCLASSINFOpdf
\else
\fi

\usepackage{hyperref}
\usepackage{graphicx}
\usepackage{amsmath}
\usepackage{bm}
\usepackage{amsfonts}
\usepackage{amssymb}
\usepackage{multirow}
\usepackage{booktabs}
\usepackage{subcaption}
\usepackage{enumerate}
\usepackage{color}
\usepackage{ifpdf}
\usepackage{ulem}
\begin{document}

\title{From Pixel to Patch: Synthesize Context-aware Features for Zero-shot Semantic Segmentation}

\author{Zhangxuan Gu, Siyuan Zhou, Li Niu*, Zihan Zhao, Liqing Zhang*
\thanks{Zhangxuan Gu is with MOE Key Lab of Artificial Intelligence, Shanghai Jiao Tong University, Shanghai,
China (email: zhangxgu@126.com).}
\thanks{Siyuan Zhou is with MOE Key Lab of Artificial Intelligence, Shanghai Jiao Tong University, Shanghai,
China (email: ssluvble@sjtu.edu.cn ).}
\thanks{Li Niu* is with MOE Key Lab of Artificial Intelligence, Shanghai Jiao Tong University, Shanghai,
China (email: ustcnewly@sjtu.edu.cn).}
\thanks{Zihan Zhao is with MOE Key Lab of Artificial Intelligence, Shanghai Jiao Tong University, Shanghai, China (email: zhao\_mengxin@sjtu.edu.cn).}
\thanks{Liqing Zhang* is with MOE Key Lab of Artificial Intelligence, Shanghai Jiao Tong University, Shanghai,
China (email: zhang-lq@cs.sjtu.edu.cn).}
\thanks{*Corresponding authors.}
}


\IEEEtitleabstractindextext{
\begin{abstract}
Zero-shot learning has been actively studied for image classification task to relieve the burden of annotating image labels. Interestingly, semantic segmentation task requires  more labor-intensive pixel-wise annotation, but zero-shot semantic segmentation has not attracted extensive research interest.
Thus, we focus on zero-shot semantic segmentation, which aims to segment unseen objects with only category-level semantic representations provided for unseen categories. In this paper, we propose a novel Context-aware feature Generation Network (CaGNet), which can synthesize context-aware pixel-wise visual features for unseen categories based on category-level semantic representations and pixel-wise contextual information. The synthesized features are used to finetune the classifier to enable segmenting unseen objects. Furthermore, we extend pixel-wise feature generation and finetuning to patch-wise feature generation and finetuning, which additionally considers inter-pixel relationship. Experimental results on Pascal-VOC, Pascal-Context, and COCO-stuff show that our method significantly outperforms the existing zero-shot semantic segmentation methods.
\end{abstract}
}

\maketitle

\IEEEdisplaynontitleabstractindextext
\IEEEpeerreviewmaketitle

\section{Introduction}\label{intro}

Zero-shot learning~\cite{lampert2013attribute} aims to deal with the problem of classifying previously unseen categories and numerous zero-shot classification methods~\cite{10.1145/2964284.2964319,10.1145/2578726.2578746,10.1145/3240508.3240715,Li2019Leveraging,Mandal2019Out,zhang2018triple,guo2017zero,rahman2018unified,huynh2020fine,yu2020episode,7883945} have been developed. Compared with the popularity of zero-shot classification, zero-shot semantic segmentation has received much less attention. 
However, semantic segmentation~\cite{long2015fully,zhao2017pyramid,chen2018deeplab,ronneberger2015u,Lin2016RefineNet,GaoLearning,fu2019stacked,bi2020polarimetric,8320527,9154612}, with the goal to classify each pixel in one image, has much higher annotation cost than the classification task. So it is in high demand to design effective zero-shot semantic segmentation methods. 

The setting of zero-shot semantic segmentation~\cite{bucher2019zero} is similar to that of zero-shot classification. All categories are divided into seen categories and unseen categories, which have no overlap. For ease of description, we refer to the entities (\emph{e.g.}, objects, pixels, features, word embeddings) belonging to seen/unseen categories as seen/unseen ones.
The training images only have pixel-wise annotations for seen categories while the test images may contain both seen and unseen objects. To enable the trained model to segment unseen objects in the testing stage, we need to transfer knowledge from seen categories to unseen categories via category-level semantic representations (\emph{e.g.}, word embedding like word2vec~\cite{mikolov2013distributed}).

\begin{figure}[tbp]
\centering
\includegraphics[width=\linewidth]{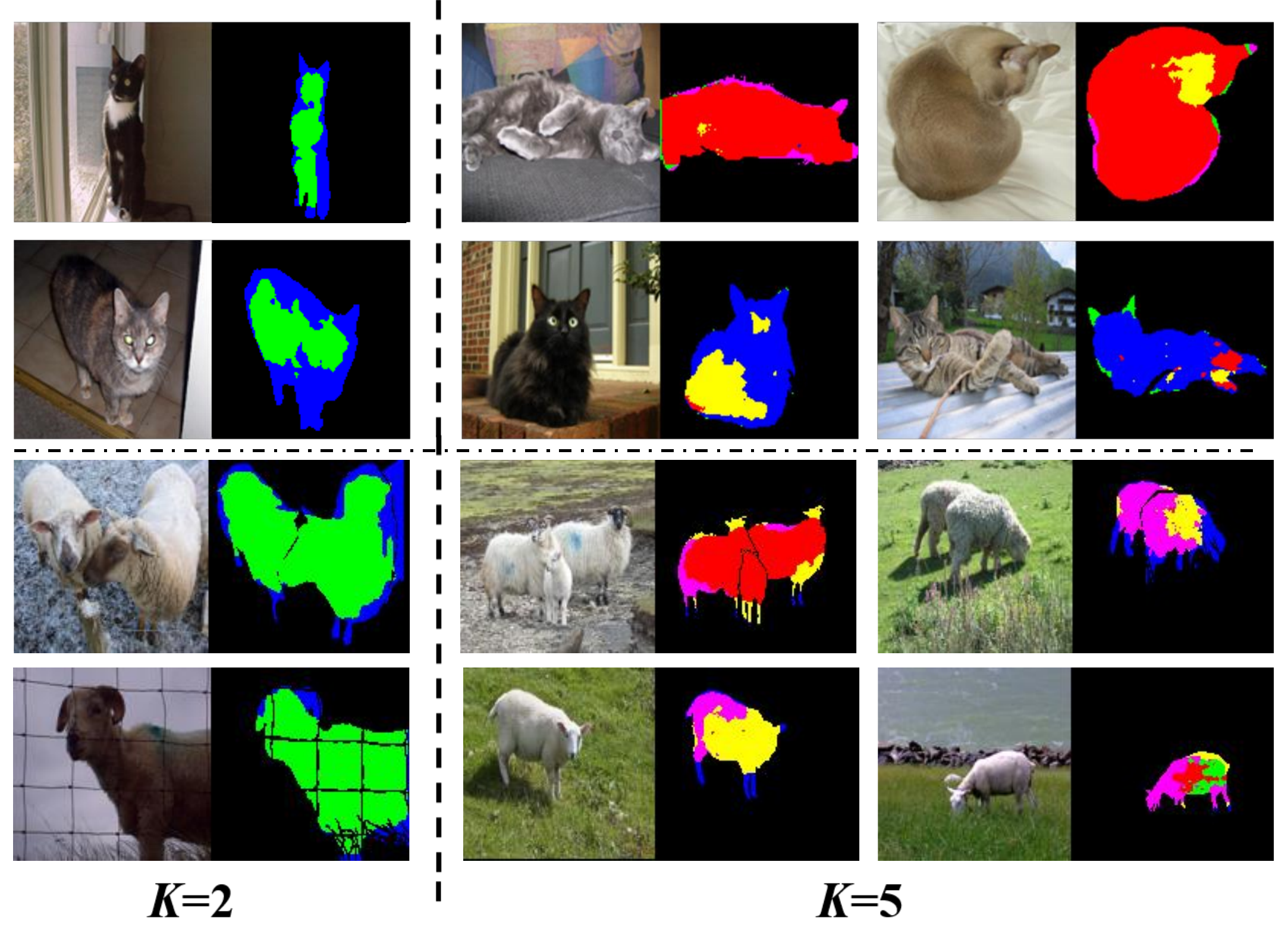}
\caption{The clustering results of pixel-wise visual features of category ``cat" and ``sheep'' with $K$ being the number of clusters ($K=2$ on the left and $K=5$ on the right). Different clusters are represented by different colors.}
\label{example}
\end{figure}

Among the existing zero-shot segmentation works methods~\cite{kato2019zero,xian2019semantic,bucher2019zero,liu2020learning,2018Attribute,lv2020learning,li2020consistent,hu2020uncertainty}, our work is more related to \cite{xian2019semantic,bucher2019zero,hu2020uncertainty,li2020consistent}, which are inductive methods aiming to segment multiple categories in an image. 
The above methods \cite{xian2019semantic,bucher2019zero,hu2020uncertainty,li2020consistent} can be divided into two groups: 1) SPNet~\cite{xian2019semantic} and Hu \emph{et al.}~\cite{hu2020uncertainty} learn a mapping from visual features to word embeddings, so that classification of all categories can be accomplished in the word embedding space; 2) ZS3Net~\cite{bucher2019zero} and CSRL~\cite{li2020consistent} learn a mapping from word embeddings to visual features, so that unseen visual features can be generated from unseen word embeddings. Our method belongs to the second group and extends ZS3Net~\cite{bucher2019zero}, so now we briefly introduce the visual feature generation pipeline used in \cite{bucher2019zero}.


Given the word embedding of one category and a random vector, ZS3Net~\cite{bucher2019zero} tends to generate pixel-wise visual features for this category in the segmentation network. The generator is trained based on seen categories, but expected to produce features for unseen categories. The generated features are used to finetune the final classifier to enable segmenting unseen objects. 
However, ZS3Net has two severe drawbacks: 1) It simply injects random vectors into the generator to generate diverse features, but the generated features may collapse into a few modes and have limited diversity, which is known as the mode collapse problem~\cite{isola2017image,mathieu2015deep,pathak2016context,NeurIPS2017_6650}; 2) It only synthesizes pixel-wise visual features, which is limited to finetuning the last $1\times 1$ convolutional layer.

\begin{figure}[tbp]
\centering
\includegraphics[width=0.9\linewidth]{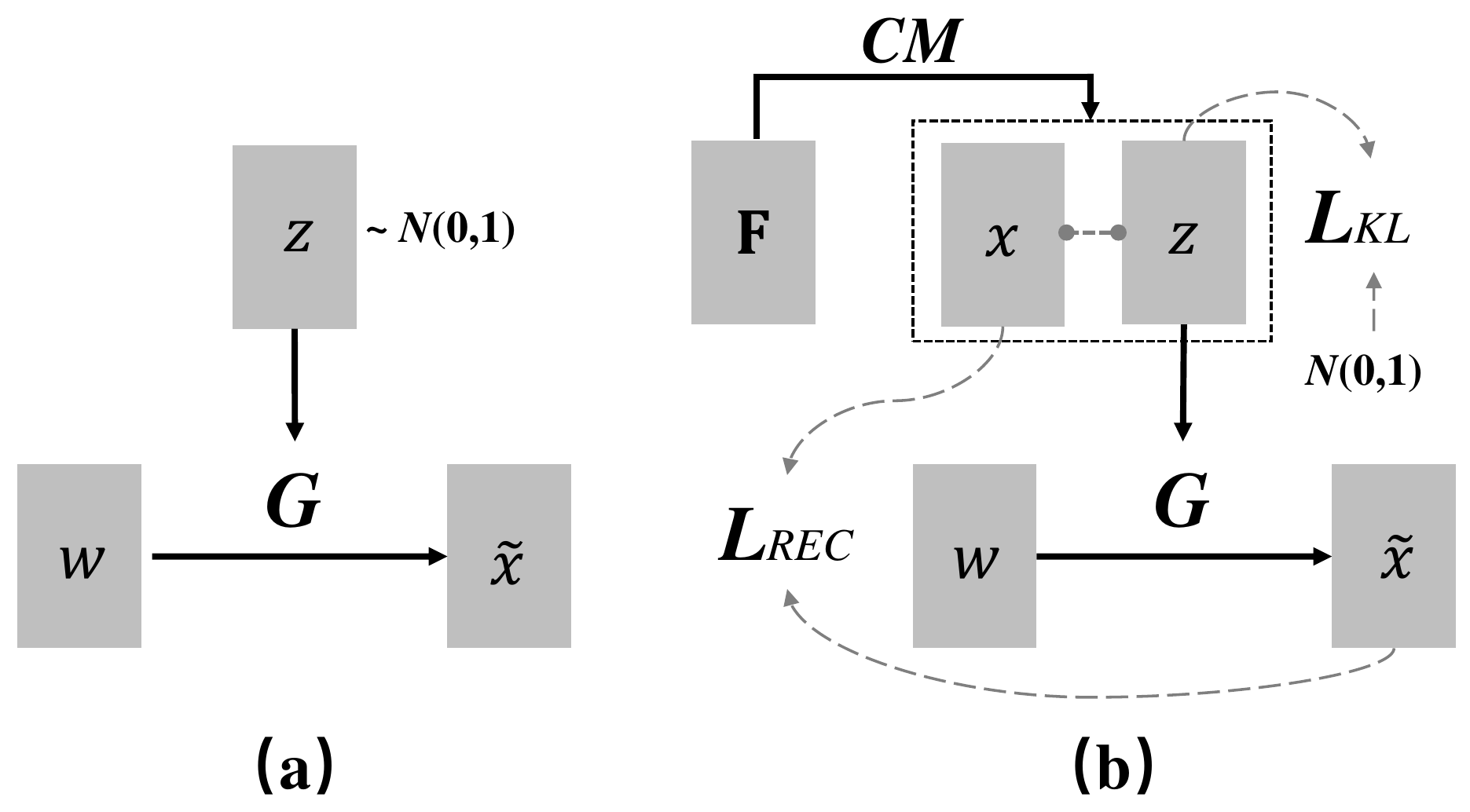}
\caption{The comparison between ZS3Net (a) and our CaGNet (b). Our contextual module $CM$ takes the output $\mathbf{F}$ of segmentation backbone as input, and outputs visual feature $\mathbf{x}$ and its corresponding contextual latent code $\mathbf{z}$ for all pixels. Our generator $G$ aims to reconstruct $\mathbf{x}$ based on $\mathbf{z}$ and word embedding $\mathbf{w}$.}
\label{diff}
\end{figure}

To overcome the first drawback of ZS3Net, one way to mitigate mode collapse is associating random vector with specific meaning and establish the one-to-one correspondence  between random vector and output~\cite{NeurIPS2017_6650}. In our task, we conjecture that pixel-wise feature highly depends on its contextual information, in which the contextual information of a pixel means the information inferred from its surrounding pixels (\emph{e.g.}, its location in the object, the pose of the object it belongs to, background objects). If so, we can associate random vector with contextual information and establish the one-to-one correspondence between pixel-wise contextual information and pixel-wise feature. 
To confirm this conjecture, we perform K-means clustering on the visual features (output from the ASPP module in Deeplabv2~\cite{chen2018deeplab}) of category ``cat" and ``sheep'' on Pascal-Context~\cite{mottaghi2014role} dataset. 
From Fig.~\ref{example}, it can be seen that pixel-wise visual features are influenced by their contextual information in a complicated and interlaced fashion. When $K=2$, the features from the exterior and interior of cat and sheep are clustered to two different groups. When $K=5$, the form of clusters is affected by adjacent or distant background objects. For instance, the red (\emph{resp.}, blue) cluster of cat features might be related to the blanket (\emph{resp.}, green plant). Similarly, the red cluster of sheep features might be related to the water. Therefore, we attempt to encode pixel-wise contextual information into contextual latent code to replace random vector, leading to our \emph{\textbf{C}}ontext-\emph{\textbf{a}}ware feature \emph{\textbf{G}}eneration \emph{\textbf{Net}}work (CaGNet).

Formally, the difference between ZS3Net and our CaGNet in feature generation is shown in Fig.~\ref{diff}. The feature generator $G$ in ZS3Net takes word embedding $\mathbf{w}$ and random vector following unit Gaussian distribution $\mathbf{z}\sim\mathcal{N}(\mathbf{0,1})$ as input to generate pixel-wise fake feature $\tilde{\mathbf{x}}$. In contrast, we use contextual module $CM$ to extract pixel-wise contextual information from feature map $\mathbf{F}$, which is encoded into contextual latent code $\mathbf{z}$.  
It is worth mentioning that our contextual module $CM$ contains a novel context selector, which can automatically decide the suitable scale of context for each pixel. 
The feature generator $G$ in our CaGNet takes in word embedding $\mathbf{w}$ and contextual latent code $\mathbf{z}$ to reconstruct real feature $\mathbf{x}$. This establishes the one-to-one correspondence between pixel-wise contextual information and pixel-wise feature, which enforces different contextual latent codes to generate different visual features.
We also enforce the distribution of $\mathbf{z}$ to approach $\mathcal{N}(\mathbf{0,1})$ to support stochastic sampling. Given one word embedding $\mathbf{w}$, our model can generate visual features of this category in different contexts by sampling  $\mathbf{z}\sim\mathcal{N}(\mathbf{0,1})$, so that the diversity of generated features can be greatly enhanced.
After training the feature generator $G$ based on seen categories, we can generate pixel-wise features for both seen and unseen categories, which are used to finetune the last $1\times1$ convolutional layer (pixel-wise finetuning) to bridge the gap between seen categories and unseen categories. 

To overcome the second drawback of ZS3Net, we extend pixel-wise feature generation to patch-wise feature generation, which can be used to finetune the classifier including convolutional layers with kernel size larger than $1$ (patch-wise finetuning). 
Patch-wise feature is synthesized on the premise of reasonable category patch, which encodes the semantic layout of a patch.
We choose PixelCNN~\cite{pixelrnn} to synthesize category patches because PixelCNN can explicitly model inter-pixel dependencies. 
Similar finetuning step for the classifier is performed based on the generated patch-wise features. Through patch-wise finetuning which considers inter-pixel relationship, our CaGNet can further improve the segmentation performance.

In summary, compared with ZS3Net, our CaGNet can produce more diverse features  with pixel-wise contextual information and support more advanced finetuning with patch-wise features. Compared with our preliminary conference version~\cite{CaGNet}, we extend pixel-wise feature generation and finetuning to patch-wise feature generation and finetuning. Our contributions can be summarized as follows,
\begin{itemize}
    \item We use pixel-wise contextual information as guidance to synthesize diverse and context-aware features for zero-shot semantic segmentation.
    \item {We integrate pixel-wise feature generation and semantic segmentation coherently in a unified network.}
    \item {We extend pixel-wise feature to patch-wise feature, by designing the corresponding patch-wise feature generation and patch-wise feature finetuning.} 
    \item Two minor contributions are 1) contextual module with a novel context selector; 2) modified PixelCNN to generate plausible category patches.
    \item Extensive experiments on three benchmark datasets demonstrate the effectiveness of our method. 
\end{itemize}

\begin{table*}[!htbp]
\centering
\begin{tabular}{c|ccc|ccc|ccc|ccc}
\toprule[1.5pt]
\multirow{2}*{Descriptions}&\multicolumn{3}{c}{One-hot Label}  &\multicolumn{3}{c}{Latent Code}   & \multicolumn{3}{c}{Word Embedding}  & \multicolumn{3}{c}{Visual Feature}       \\ 
~    & map  & pixel  & patch  & map  & pixel  & patch & map  & pixel  & patch & map  & pixel  & patch \\ \hline
Real&$\mathbf{Y}_n^s$& $\mathbf{y}^s_{n,i}$&$\mathbf{by}_{n,i}^s$&$\mathbf{Z}_n$&  $\mathbf{z}_{n,i}$&$\mathbf{bz}_{n,i}$&$\mathbf{W}_n^s$&$\mathbf{w}_{n,i}^s$&  $\mathbf{bw}^s_{n,i}$&$\mathbf{X}_n^s$&  $\mathbf{x}^s_{n,i}$&$\mathbf{bx}_{n,i}^s$ \\
Fake&${\tilde{\mathbf{Y}}}^{s\cup u}_m$&$\tilde{\mathbf{y}}^{s\cup u}_{m,i}$&${\widetilde{\mathbf{by}}}_{m,i}^{s\cup u}$& ${\tilde{\mathbf{Z}}}_m$&$\tilde{\mathbf{z}}_{m,i}$&${\widetilde{\mathbf{bz}}}_{m,i}$& ${\tilde{\mathbf{W}}}^{s\cup u}_m$&$\tilde{\mathbf{w}}^{s\cup u}_{m,i}$ & ${\widetilde{\mathbf{bw}}}_{m,i}^{s\cup u}$ &${\tilde{\mathbf{X}}}^{s\cup u}_m$  & $\tilde{\mathbf{x}}^{s\cup u}_{m,i}$& ${\widetilde{\mathbf{bx}}}_{m,i}^{s\cup u}$\\ \bottomrule[1.5pt]
\end{tabular}
\caption{Notation of main symbols used in this manuscript. We use upperscript $s$ (\emph{resp.}, $u$) to indicate seen (\emph{resp.}, unseen) categories. The subscripts $n$ (\emph{resp.}, $m$) denote the $n$-th (\emph{resp.}, $m$-th) real (\emph{resp.}, fake) map. The subscript $i$ denotes the $i$-th pixel/patch in the corresponding map.}
\label{notations}
\end{table*}

\section{Related Works}

\noindent\textbf{Semantic Segmentation:} Semantic segmentation is a long-standing and challenging problem in computer vision. Many methods like~\cite{zhao2017pyramid,chen2018deeplab,ronneberger2015u,Lin2016RefineNet,GaoLearning,8320527,9154612} have made great progress in utilizing rich contextual information, because the category predictions of target objects are often influenced by nearby objects or background scenes.
For example, PSPNet~\cite{zhao2017pyramid} and Deeplab~\cite{chen2018deeplab} designed specialized pooling layers to fuse the contextual information from feature maps of different scales. U-Net~\cite{ronneberger2015u} and RefineNet~\cite{Lin2016RefineNet} focused on designing network architectures that better combine low-level features and high-level ones for ampler contextual cues.
The above works motivate us to incorporate contexts into feature generation. However, all the above standard semantic segmentation models require heavy pixel-wise annotations of all categories during training, so they cannot be applied to the zero-shot segmentation task. 

\noindent\textbf{Weakly Supervised Semantic Segmentation:} Because semantic segmentation task heavily relies on dense pixel-wise annotations,
there has been an increasing interest in weakly supervised methods like image-level~\cite{oh2017exploiting,papandreou2015weakly,Yao2015Semantic}, box-level~\cite{khoreva2017simple,GraphNet}, or scribble-level~\cite{lin2016scribblesup} semantic segmentation, to relieve the burden of pixel-wise annotations. In this paper, we take a step further and only provide category-level semantic representations for unseen categories, following the zero-shot semantic segmentation setting~\cite{xian2019semantic,bucher2019zero}.

\begin{figure*}[tbp]
\centering
\includegraphics[width=\linewidth]{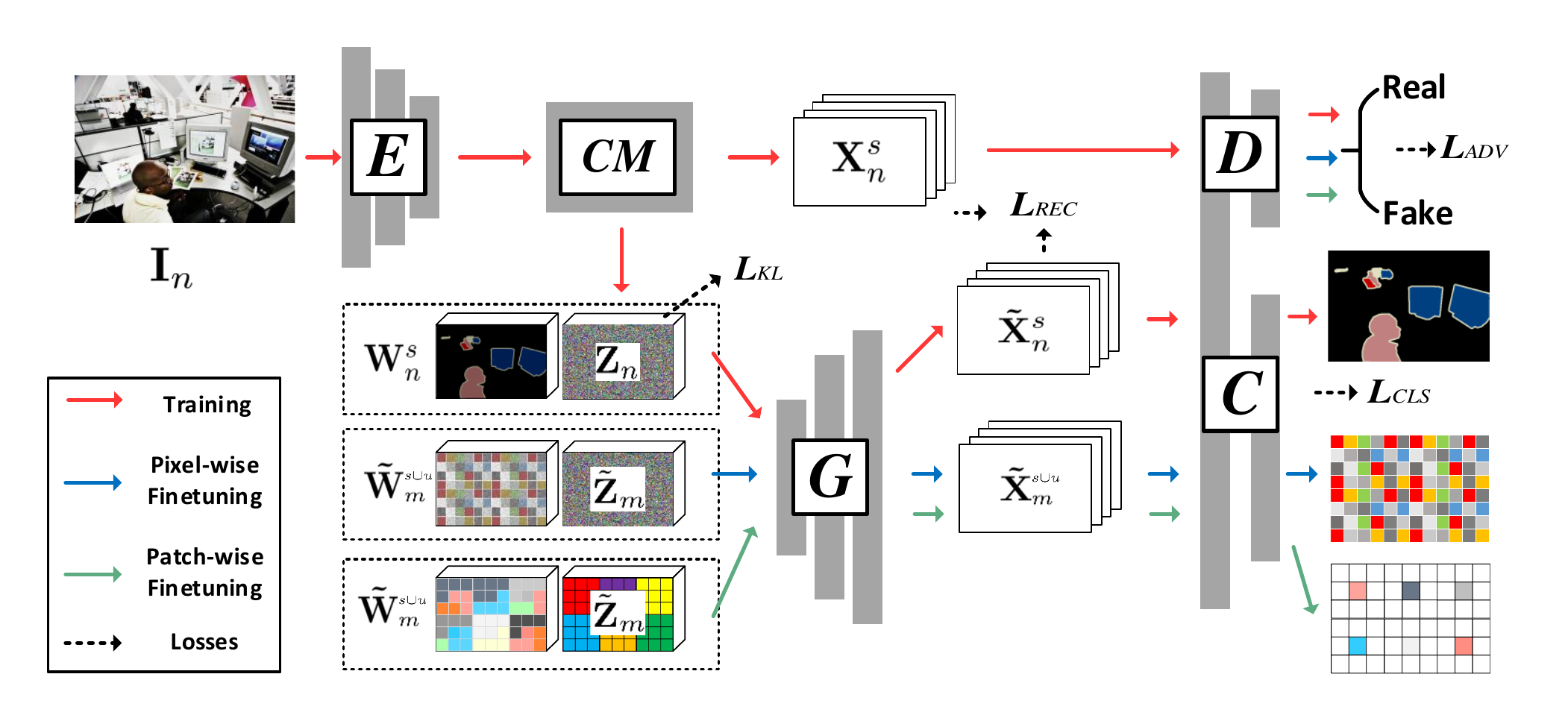}
\caption{\textbf{Overview of our method.} Our model contains segmentation backbone $E$, contextual module $CM$, feature generator $G$, discriminator $D$, and classifier $C$. $\mathbf{W}$, $\mathbf{Z}$, and $\mathbf{X}$ represent word embedding map, contextual latent code map, and feature map respectively (see Sec.~\ref{sec:CM} and \ref{sec:context_feature_generator} for detailed definition). Optimization steps are separated into training step and finetuning step indicated by different colors (see Sec.~\ref{sec:optimization}).}
\label{ours}
\end{figure*}


\noindent\textbf{Zero-shot Learning: }Zero-shot learning (ZSL) was first introduced by~\cite{Lampert2009Learning} in classification task. ZSL stands for the setting in which all training data are from seen categories while test data may come from unseen categories. The key of ZSL is to transfer knowledge from seen categories to unseen ones via category-level semantic representations. Many existing ZSL methods~\cite{akata2015label,frome2013devise,romera2015embarrassingly,fu2015transductive,10.1145/2964284.2964319,10.1145/2578726.2578746,10.1145/3240508.3240715,7883945} attempted to learn a mapping between feature space and semantic space. In some cases, we have one sample for each unseen category, which is referred to as one-shot learning. One-shot learning has been studied in many fields like~\cite{liu2020iterative,shaban2017one,zhang2020sg,zhang2019pyramid,caelles2017one}.

Recently, training a conditional generator to synthesize features for unseen categories is a popular approach in zero-shot classification. For example, the method in~\cite{xian2018feature} first generated features using word embeddings and random vectors, which was further improved by later works \cite{FelixMulti,xian2019f,Mert2019Gradient,Li2019Leveraging,Mandal2019Out}. However, these methods generated visual features without involving any contextual information. In contrast, due to the uniqueness of semantic segmentation task, we utilize contextual information to facilitate feature generation.

\noindent\textbf{Generative Adversarial Networks:} Generative adversarial network~\cite{goodfellow2014generative} (GAN) usually consists of two models: the generator that is trained to generate new samples and the discriminator that attempts to classify samples as either real or fake. GAN has lead to considerable progress in many real-world applications~\cite{brock2018large,isola2017image,ledig2017photo,MaoLeast,Mert2019Gradient}. To name a few, for unconditional image generation, BigGAN~\cite{brock2018large} and LS-GAN~\cite{MaoLeast} aim to generate realistic photographs. For conditional image generation, pix2pix~\cite{isola2017image} is designed for many image-to-image translation tasks. For cross-modal generation, StackGAN~\cite{zhang2017stackgan} is proposed to generate realistic-looking photographs from textual descriptions of simple objects like birds and flowers. GAN has also been widely used in low-level computer vision tasks. For example, for image inpainting, GAIN~\cite{zhang2019gain} describes the use of GAN to perform photograph inpainting or hole filling. For super-resolution, SRGAN~\cite{ledig2017photo} has been introduced to generate output images with much higher pixel resolution. In Zero-Shot Learning (ZSL), many researchers utilize GAN to generate features for image classification~\cite{xian2018feature,Mert2019Gradient,Li2019Leveraging}. Inspired by their success, we propose a GAN with contextual information to synthesize unseen features for zero-shot semantic segmentation.

\noindent\textbf{Zero-shot Semantic Segmentation: }To the best of our knowledge, zero-shot  segmentation appeared in some prior works~\cite{kato2019zero,xian2019semantic,bucher2019zero,liu2020learning,2018Attribute,lv2020learning,li2020consistent,hu2020uncertainty}. Among them, transductive methods \cite{liu2020learning,lv2020learning} require test images during training, while this paper focuses on inductive zero-shot semantic segmentation without requiring test images during training. The methods \cite{kato2019zero,2018Attribute} only handle binary segmentation task, while this paper focuses on general semantic segmentation. 

The remaining works \cite{xian2019semantic,hu2020uncertainty,bucher2019zero,li2020consistent} are closely related to our work and representative for two groups of methods: 1) mapping from visual features to semantic representations and 2) mapping from semantic representations to visual features. 
In the first group, SPNet~\cite{xian2019semantic} transferred knowledge between seen and unseen categories via semantic projection layer, which projects features to semantic word embeddings. Another method \cite{hu2020uncertainty} learned the visual-semantic mappings with global information to automatically discriminate between representative samples and noisy ones during training. 
In the second group, ZS3Net~\cite{bucher2019zero} used semantic word embeddings to generate pixel-wise unseen features, which are utilized to finetune a classifier trained on seen categories. Moreover, to capture spatial object arrangement, ZS3Net is extended to ZS3Net(GC) by utilizing Graph Convolutional Network (GCN)~\cite{kipf2016semi-supervised} to represent each image as an object relational graph. However, ZS3Net(GC) only considers object-level contextual information and the object relational graphs involving unseen categories are usually unavailable. More recently, Li \emph{et al.} \cite{li2020consistent} proposed a Consistent Structural Relation Learning (CSRL) approach to constrain the generation of unseen visual features by exploiting the structural relations between seen and unseen categories.
Our method belongs to the second group and holds several advantages: 1) we leverage rich contextual information to guide feature generation; 2) we can generate both pixel-wise features and patch-wise features. Compared with ZS3Net(GC), our method does not require object relational graphs involving unseen categories.

\section{Methodology}\label{method}


%

In zero-shot semantic segmentation task, all categories are split into seen categories $\mathcal{C}^s$ and unseen categories $\mathcal{C}^u$, where $\mathcal{C}^s\cap \mathcal{C}^u=\emptyset$.
Training images only have pixel-wise annotations of $\mathcal{C}^s$, while test images contain objects of $\mathcal{C}^s\cup \mathcal{C}^u$.
To fulfil the knowledge transfer across categories, we resort to category-level word embeddings $\{{{\mathbf{w}}}_c|c\in \mathcal{C}^s\cup \mathcal{C}^u\}$, in which ${{\mathbf{w}}}_c \in \mathcal{R}^{d}$ is the word embedding of category $c$.

\subsection{Overview}\label{overall}

Generally speaking, any segmentation network can be split into a backbone $E$ and a classifier $C$. $E$ extracts the feature map of an input image and $C$ outputs the final segmentation results based on this feature map. Our CaGNet can be built upon an arbitrary segmentation network. 
In this paper, we adopt Deeplabv2~\cite{chen2018deeplab} due to its remarkable performance in semantic segmentation.
A standard Deeplabv2 network trained on seen categories is unable to segment unseen objects. We attempt to learn a feature generator $G$ based on seen categories, which  can synthesize fake features for unseen categories. The classifier in Deeplabv2 is finetuned with the synthesized fake features,  and thus able to segment unseen objects.  

Fig.~\ref{ours} shows the overall architecture of our CaGNet. We add our proposed contextual module $CM$ between the backbone $E$ and the classifier $C$. $CM$ can produce pixel-wise contextual latent code, which  guides the generator $G$ to generate more diverse and context-aware features. The backbone $E$ and the contextual module $CM$ output real features from the input image. The generator $G$ outputs fake features from the contextual latent code and the word embedding. Real features and fake features are then delivered to the classifier $ C $ and the discriminator $ D $, which could share partial layers to reduce the number of model parameters~\cite{xu2018pad-net:}.  Finally, $C$ outputs segmentation results and $ D $ discriminates real features from fake ones. As shown in Fig.~\ref{ours}, the contextual module $CM$ and the classifier $C$ link the segmentation network $\{E, CM, C\}$ and the feature generation network $\{CM, G, D, C\}$.

Next, we will detail our contextual module in Sec.~\ref{sec:CM} and pixel-wise (\emph{resp.}, patch-wise) feature generator in Sec.~\ref{sec:context_feature_generator} (\emph{resp.}, \ref{sec:pixelcnn++}). For better representation, we use capital letter in bold (\emph{e.g.}, $\mathbf{X}$) to denote a map and small letter in bold (\emph{e.g.}, $\mathbf{x}_i$) to denote its pixel-wise vectors. Besides, we use upperscript $s$ (\emph{resp.}, $u$) to represent seen (\emph{resp.}, unseen) categories. The main symbols used in this paper are summarized in Table~\ref{notations}.

\subsection{Contextual Module}\label{sec:CM}

\begin{figure}[tbp]
\centering
\includegraphics[width=\linewidth]{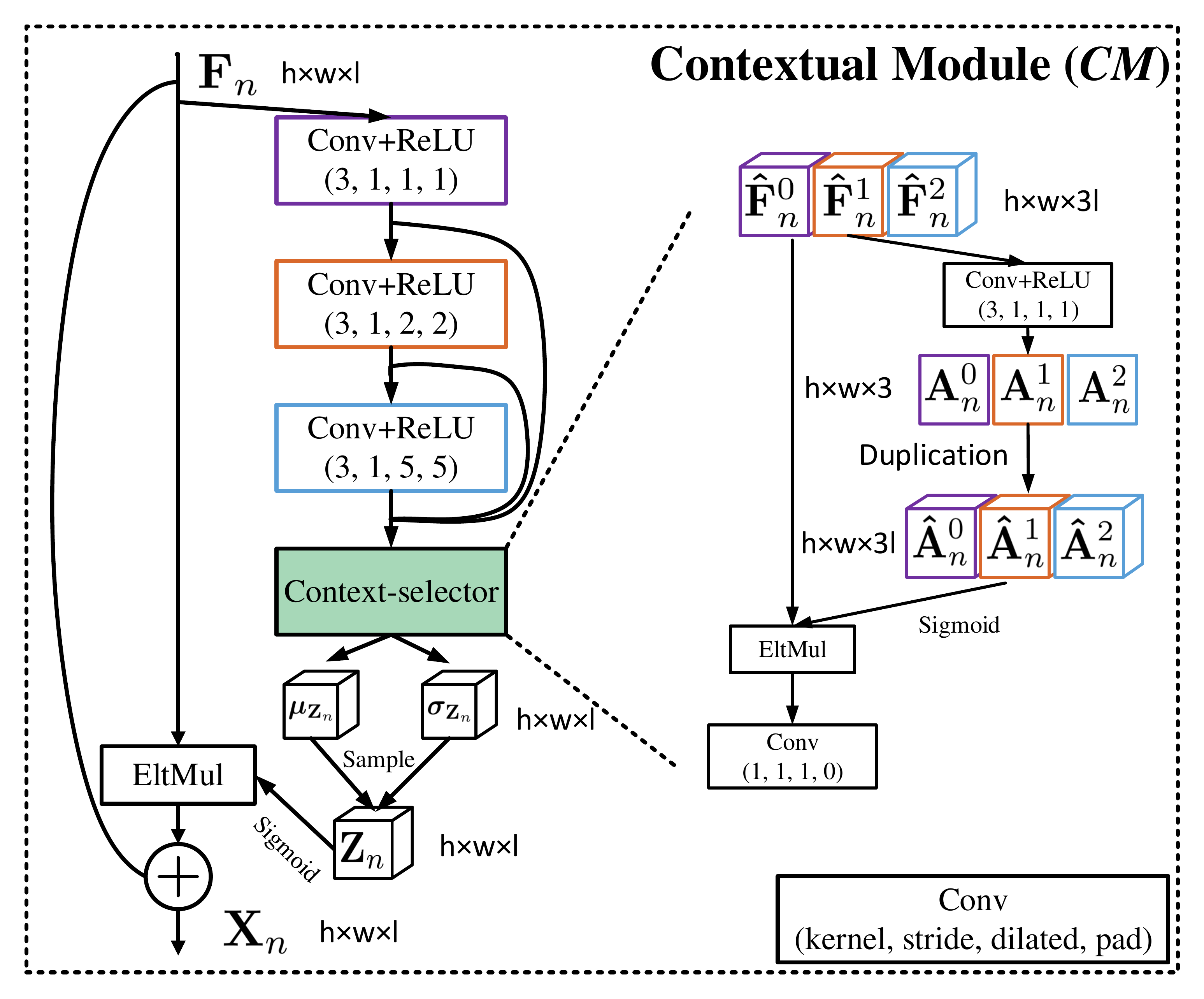}
\caption{\textbf{Contextual Module.} We aggregate the contextual information of different scales using our context selector. Then, the aggregated contextual information produces latent distribution for sampling contextual latent code. }
\label{contextual}
\end{figure}

Pixel-wise contextual information of a pixel is the aggregated information of its neighbouring pixels. We design a contextual module $CM$, which is the core module of our CaGNet, to gather pixel-wise contextual information for each pixel on the output feature map from segmentation backbone $E$. For the $n$-th image, we use $\mathbf{F}_n\in \mathcal{R}^{h\times w\times l}$ to denote the output feature map from $E$. Our contextual module $CM$ is directly applied to $\mathbf{F}_n$ as illustrated in Fig.~\ref{ours}.

1) \textbf{Context Map and Selector:} In our $CM$, we apply several dilated convolutional (conv) layers~\cite{Yu2015Multi} to $\mathbf{F}_n$ to obtain three context maps ${\hat{\mathbf{F}}}^0_n,{\hat{\mathbf{F}}}^1_n,{\hat{\mathbf{F}}}^2_n\in \mathcal{R}^{h\times w\times l}$, which capture the contextual information of three different scales.
Besides, we conjecture that the visual features of some pixels may be mainly affected by small-scale contextual information (\emph{e.g.}, the pose or inner parts of objects), while others may be mainly affected by large-scale contextual information (\emph{e.g.}, distant background objects). Hence, we propose a context selector to learn different scale weights for different pixels adaptively.
Particularly, we learn a $3$-channel scale weight map $\mathbf{{A}}_n=[\mathbf{A}_n^0,\mathbf{A}_n^1,\mathbf{A}_n^2]\in \mathcal{R}^{h\times w\times 3}$, where $\mathbf{A}_n^t$ contains the weights of the $t$-th scale.  Then, we duplicate each $\mathbf{A}_n^t$ to ${\hat{\mathbf{A}}}_n^t \in \mathcal{R}^{h\times w\times l}$, and obtain the weighted concatenation of context maps $[{\hat{\mathbf{F}}}^0_n\odot{\hat{\mathbf{A}}}_n^0,{\hat{\mathbf{F}}}^1_n\odot{\hat{\mathbf{A}}}_n^1,{\hat{\mathbf{F}}}^2_n\odot{\hat{\mathbf{A}}}_n^2]\in \mathcal{R}^{h\times w\times 3l}$ with $\odot$ indicating elementwise-multiplication. More details can be found in our conference version~\cite{CaGNet}. 

2) \textbf{Contextual Latent Code:} As discussed in Sec.~\ref{intro}, contextual latent code, which is assumed to encode pixel-wise contextual information, functions as both the output of contextual module $CM$ and the input of feature generator $G$. Intuitively, given a pixel in a cat near a tree, its contextual latent code may encode its relative location in the cat, its nearby region in the cat,  background objects like the tree, \emph{etc}.

Next, we will describe how to calculate contextual latent code based on context maps. First, we apply a $1\times 1$ conv layer to $[{\hat{\mathbf{F}}}^0_n\odot{\hat{\mathbf{A}}}_n^0,{\hat{\mathbf{F}}}^1_n\odot{\hat{\mathbf{A}}}_n^1,{\hat{\mathbf{F}}}^2_n\odot{\hat{\mathbf{A}}}_n^2]$ to obtain $\bm{\mu}_{\mathbf{Z}_n}\in \mathcal{R}^{h\times w\times l}$ with pixel-wise $\bm{\mu}_{\mathbf{z}_{n,i}}$ and $\bm{\sigma}_{\mathbf{Z}_n}\in \mathcal{R}^{h\times w\times l}$ with pixel-wise $\bm{\sigma}_{\mathbf{z}_{n,i}}$.
Following~\cite{goodfellow2014generative}, we sample pixel-wise contextual latent code $\mathbf{z}_{n,i}$ from Gaussian distribution $\mathcal{N}(\bm{\mu}_{\mathbf{z}_{n,i}},\bm{\sigma}_{\mathbf{z}_{n,i}})$ by using $\mathbf{z}_{n,i}=\bm{\mu}_{\mathbf{z}_{n,i}}+\epsilon \bm{\sigma}_{\mathbf{z}_{n,i}} \in \mathcal{R}^{l}$, where $\epsilon\sim\mathcal{N}(0,1)$.
We also adopt a KL divergence loss to enforce $\mathcal{N}(\bm{\mu}_{\mathbf{z}_{n,i}},\bm{\sigma}_{\mathbf{z}_{n,i}})$ to approach $\mathcal{N}(\mathbf{0,1})$: 
\begin{eqnarray}
\mathcal{L}_{KL} = \mathcal{D}_{KL}[\mathcal{N}(\bm{\mu}_{\mathbf{z}_{n,i}},\bm{\sigma}_{\mathbf{z}_{n,i}})||\mathcal{N}(\mathbf{0,1})].
\end{eqnarray}

In this way, our contextual latent code supports stochastic sampling to generate visual features, by sampling $\mathbf{z}_{n,i}$ from $\mathcal{N}(\mathbf{0,1})$.

Contextual latent code map $\mathbf{Z}_n \in \mathcal{R}^{h\times w\times l}$ is acquired by aggregating all $\mathbf{z}_{n,i}$ for the $n$-th image. The main function of $\mathbf{Z}_n$ is to facilitate feature generation, but it can also be utilized to enhance feature map $\mathbf{F}_n$. Inspired by~\cite{hu2018gather}, we use $\mathbf{Z}_n$ as residual attention to obtain a new feature map $\mathbf{X}_n = \mathbf{F}_n+\mathbf{F}_n\odot \phi(\mathbf{Z}_n) \in \mathcal{R}^{h\times w\times l}$, where $\phi$ denotes sigmoid function. Compared with $\mathbf{F}_n$, new feature map $\mathbf{X}_n$ can bring a little improvement for the segmentation results as mentioned in~\cite{CaGNet}. Note that we use $\mathbf{X}_n$ instead of $\mathbf{F}_n$ as the target real features for feature generation (see Sec.~\ref{sec:context_feature_generator}).

\subsection{Pixel-wise Feature Generation}\label{sec:context_feature_generator}

In this section, we first introduce how to train feature generator $G$ based on training images. Note that we only utilize the seen pixels while ignoring the unseen pixels in the training images.
When processing the $n$-th input image $\mathbf{I}_n$, our $CM$ has two outputs: contextual latent code map $\mathbf{Z}_n \in \mathcal{R}^{h\times w\times l}$ with pixel-wise $\mathbf{z}_{n,i}$ and real visual feature map $\mathbf{X}_n^s$ with pixel-wise  $\mathbf{x}_{n,i}^s$. We follow~\cite{bucher2019zero} to use the down-sampled segmentation label map $\mathbf{Y}_n^s \in \mathcal{R}^{h\times w\times (|\mathcal{C}^s|+|\mathcal{C}^u|)}$ that has the same spatial resolution as $\mathbf{X}_n^s$. In detail, $c_{n,i}^s$ denotes the category label of the $i$-th pixel on $\mathbf{X}_n^s$. $\mathbf{y}_{n,i}^s$ (the $i$-th pixel-wise vector in $\mathbf{Y}_n^s$) is the one-hot label vector corresponding to $c_{n,i}^s$.
The word embedding of the $i$-th pixel is $\mathbf{w}^s_{n,i} = {{\mathbf{w}}}_{c_{n,i}^s}$ corresponding to its category label $c_{n,i}^s$.

Then, the generator $G$ takes $\mathbf{z}_{n,i}$ and $\mathbf{w}_{n,i}^s$ as input to synthesize pixel-wise fake feature ${\mathbf{\tilde{x}}}_{n,i}^s=G(\mathbf{z}_{n,i},\mathbf{w}_{n,i}^s)$.
As discussed in Sec.~\ref{intro}, we expect to reconstruct pixel-wise real feature $\mathbf{x}^s_{n,i}$ based on  $\mathbf{z}_{n,i}$ and $\mathbf{w}_{n,i}^s$, with the assumption of one-to-one correspondence between pixel-wise feature and pixel-wise contextual information for a specific category. Therefore, we apply an L2 reconstruction loss $\mathcal{L}_{REC}$:
\begin{eqnarray}
\mathcal{L}_{REC}=\sum_{n,i}||\mathbf{x}^s_{n,i}-{\tilde{\mathbf{x}}}^s_{n,i}||^2_2.
\end{eqnarray}

The real features $\mathbf{x}^s_{n,i}$ and fake features $\tilde{\mathbf{x}}^s_{n,i}$ are fed into classifier $C$ and discriminator $D$. 
Similar to most segmentation methods~\cite{long2015fully,zhao2017pyramid,chen2018deeplab,Lin2016RefineNet}, we train the classifier $C$ using cross-entropy classification loss:
\begin{eqnarray}\label{eqn:L_cls}
\mathcal{L}_{CLS} =-\sum_{n,i}\mathbf{y}_{n,i}^s\log(C(\mathbf{x}_{n,i}^s)).
\end{eqnarray}
To make the generated fake features indistinguishable from real ones, 
we adopt the adversarial loss $\mathcal{L}_{ADV}$ following~\cite{MaoLeast}:
\begin{eqnarray}\label{eqn:L_adv}
\mathcal{L}_{ADV} =\sum_{n,i} (D(\mathbf{x}_{n,i}^s))^2 + (1-D({\tilde{\mathbf{x}}}_{n,i}^s))^2,
\end{eqnarray}
in which sigmoid function is used to normalize the discriminator scores from $D$ into $[0,1]$, with $1$ indicating real features and $0$ indicating fake ones.

Although our generator $G$ is trained on seen categories, it is able to synthesize pixel-wise features for both seen and unseen categories. By randomly sampling latent code $\mathbf{z}\sim\mathcal{N}(0,1)$ to replace the contextual latent code $\mathbf{z}_{n,i}$, we can synthesize fake feature $G(\mathbf{z}, {{\mathbf{w}}}_c)$ for an arbitrary category $c\in \mathcal{C}^s\cup \mathcal{C}^u$. Intuitively, $G(\mathbf{z}, {{\mathbf{w}}}_c)$ stands for the pixel-wise feature of category $c$ in the context specified by $\mathbf{z}$. The synthesized features will be used to finetune the classifier $C$ and enable segmenting unseen objects (see Sec.~\ref{sec:optimization}).

\subsection{Patch-wise Feature Generation}\label{sec:pixelcnn++}

\begin{figure*}[tbp]
\centering
\includegraphics[width=\linewidth]{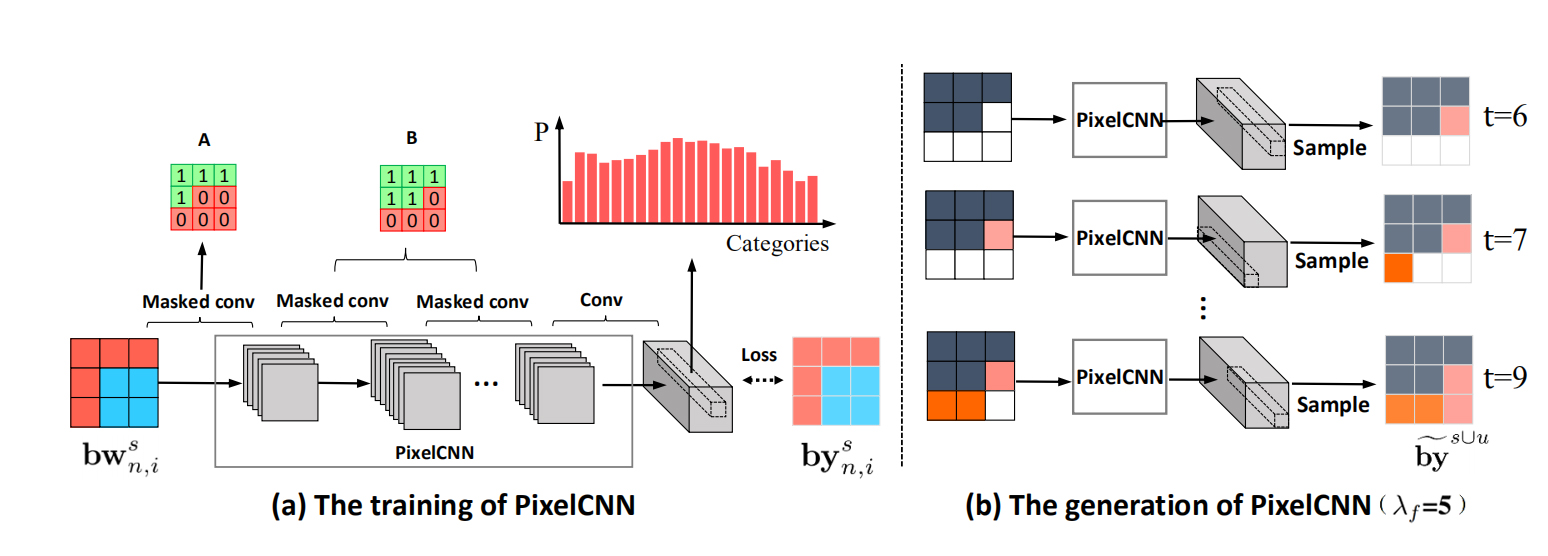}
\caption{The visualization of the training (a) and generation (b) procedure of PixelCNN when patch size $k=3$. In (a), the category distribution of $k\times k$ pixels are predicted in parallel by using masked conv layers (type A or B). In (b), the category distribution of each pixel is predicted sequentially. Different colors indicate different categories.}
\label{pixelcnn}
\end{figure*}

By using synthetic pixel-wise features, we can only finetune $1\times 1$ convolutional (conv) layer without considering the inter-pixel relationship. If we can generate patch-wise features (\emph{e.g.}, $3\times3$ patch), we can finetune the classifier formed by conv layers with larger kernel size (\emph{e.g.}, $3\times 3$ kernel), which can be better adapted to unseen categories by considering inter-pixel relationship. In this paper, we synthesize small patches ($3\times 3$) rather than large patches because generating large category patches is very challenging considering the shapes, poses, or locations of multiple objects, especially for unseen objects.

From a naive point of view, we can randomly stack pixel-wise features to construct patch-wise features. However, the semantic layout, \emph{i.e.}, the categories of pixels within a patch, may be unreasonable. For example, it can be rarely seen in an image that the ground is above the sky. To pursue plausible semantic layout, we need to generate reasonable patch-wise category labels (\emph{i.e.} category patch) before generating patch-wise features (\emph{i.e.} feature patch).
Among typical generative models like GAN~\cite{goodfellow2014generative}, VAE~\cite{kingma2013auto}, and PixelCNN~\cite{pixelrnn}, we choose PixelCNN to synthesize category patches because it can explicitly model the inter-pixel dependencies as we expect. 
In particular, PixelCNN learns plausible semantic layout based on word embeddings, and such knowledge can be transferred from seen categories to unseen categories.
Then, with the generated category patches containing both seen and unseen categories, we can synthesize feature patches.
In this process, we use the same contextual latent code for all pixels in a patch, because the contextual information of neighboring pixels in a small patch should be similar. Finally, the synthesized feature patches are used to finetune the classifier.

1) \textbf{Overview of PixelCNN:} In this section, we introduce PixelCNN~\cite{pixelrnn} (denoted as $P$), a popular auto-regressive generative model which can capture the inter-pixel dependencies. 
We creatively modify PixelCNN to generate reasonable category patches instead of natural images as it was conventionally developed for. 
Conventional PixelCNN~\cite{pixelrnn} scans an image row by row and pixel by pixel. For each pixel, it predicts the distribution of its pixel value (RGB values) conditioned on the pixel values of observed pixels.
Thus, conventional PixelCNN takes the observed pixel values as input to predict the remaining pixel values. 

In our designed PixelCNN, we take the word embeddings of the observed pixels as input to predict the category distributions of the remaining pixels. 
By virtue of category-level word embeddings, the knowledge of semantic layout can be transferred from seen categories to unseen categories.
For ease of representation, we flatten one $k\times k$ category patch as a sequence $\mathbf{c}=(c_1,c_2,\cdots,c_{k^2})$, where these $k^2$ pixels are taken from the patch row by row. The category distribution of $t$-th pixel in the category patch depends on the word embeddings of previous observed pixels, \emph{i.e.}, $p({c_t}|\mathbf{w}_{c_1},\mathbf{w}_{c_2},\cdots,\mathbf{w}_{c_{t-1}})$. Following~\cite{pixelrnn}, the estimated logarithmic joint distribution of the category patch $\log p(\mathbf{c})$ can be written as the summation of logarithmic conditional distributions over all pixels in the patch:
\begin{eqnarray}\label{pc}
\log p(\mathbf{c}) = \sum_{t=1}^{k^2} \log p({c_t}|\mathbf{w}_{c_1},\mathbf{w}_{c_2},\cdots,\mathbf{w}_{c_{t-1}}).
\end{eqnarray}
Both the training stage and the generation stage of PixelCNN are related to (\ref{pc}), which will be detailed later.

2) \textbf{Training PixelCNN:} We aim to maximize (\ref{pc}) when training PixelCNN with seen categories, so that the knowledge of semantic layout in the training patches can be learnt. To model the conditional probability $p({c_t}|\mathbf{w}_{c_1},\mathbf{w}_{c_2},\cdots,\mathbf{w}_{c_{t-1}})$ in (\ref{pc}), PixelCNN utilizes $3\times 3$ masked convolutional layers to avoid seeing the future pixels, as shown in Fig.~\ref{pixelcnn}. Following \cite{pixelrnn}, we use two types of masks named mask A and mask B. The only difference between them lies in whether the central entry is $0$ or $1$. Mask A is only applied to the first layer and mask B is applied to all the subsequent layers. The first layer uses Mask A to exclude the word embedding of the central pixel when predicting the category distribution of this pixel.
More details of masked convolutional layers can be found in \cite{pixelrnn}. By stacking one mask A layer, several mask B layers, and a $1\times 1$ conv layer as PixelCNN (see Fig.~\ref{pixelcnn}), the output category distribution of each pixel will depend on the word embeddings of all previous pixels, which coincides with (\ref{pc}).

To obtain the training patches to train PixelCNN, we slice the label map $\mathbf{Y}_n^s$ into real category patches $\mathbf{by}^s_{n,i}\in \mathcal{R}^{k\times k\times (|\mathcal{C}^s|+|\mathcal{C}^u|)}$. Note that we discard those patches containing unseen pixels.
The corresponding word embedding patches are $\mathbf{bw}^s_{n,i}\in \mathcal{R}^{k\times k\times d}$. Taking $\mathbf{bw}^s_{n,i}$ as input, $P$ outputs the normalized category distribution of each pixel, \emph{i.e.}, $P(\mathbf{bw}^s_{n,i})\in \mathcal{R}^{k\times k\times (|\mathcal{C}^s|+|\mathcal{C}^u|)}$. After flattening $\mathbf{by}^s_{n,i}$ and $P(\mathbf{bw}^s_{n,i})$ into one-dim vectors, the training loss of PixelCNN can be represented by the cross-entropy classification loss:
\begin{eqnarray}\label{Ploss}
\mathcal{L}_{P}=-\sum_{n,i}{\mathbf{by}^s_{n,i}\log(P(\mathbf{bw}^s_{n,i}))}.
\end{eqnarray}
We optimize $P$ by minimizing $\mathcal{L}_{P}$, which is equivalent to maximizing (\ref{pc}). 
Note that $P$ is optimized independently from the network modules in Fig.~\ref{ours}.

3) \textbf{Category patch generation by PixelCNN:} Based on the trained model $P$, we can generate category patches containing both seen and unseen categories in the form of $\widetilde{\mathbf{by}}^{s\cup u}\in \mathcal{R}^{k\times k\times (|\mathcal{C}^s|+|\mathcal{C}^u|)}$. As shown in Fig.~\ref{pixelcnn}, the training process is paralleled while the generation process is sequential, that being said, we generate each patch row-by-row and pixel-by-pixel in $k^2$ steps.
In step $t\in\{1,2,\cdots,k^2\}$, the trained PixelCNN model $P$ takes an incomplete word embedding patch filled with the word embeddings of previous $t-1$ pixels as input and outputs the normalized category distribution of the $t$-th pixel, which is denoted as $\mathbf{p}_{t}=p({c_t}|\mathbf{w}_{c_1},\mathbf{w}_{c_2},\cdots,\mathbf{w}_{c_{t-1}}) \in \mathcal{R}^{(|\mathcal{C}^s|+|\mathcal{C}^u|)}$.
The category label of the current $t$-th pixel $\widetilde{c}^{s\cup u}_t$ should be sampled from $\mathbf{p}_{t}$, \emph{i.e.}, $\widetilde{c}^{s\cup u}_t\sim\mathbf{p}_{t}$. We use $\mathbf{y}^{s\cup u}_t\in \mathcal{R}^{(|\mathcal{C}^s|+|\mathcal{C}^u|)}$ to represent the one-hot label vector of $\widetilde{c}^{s\cup u}_t$ and fill this $\mathbf{y}^{s\cup u}_t$ into the $t$-th pixel of the incomplete category patch. 
On the whole, we repeat the above procedure for $t=1,2,\cdots,k^2$ until $t$ reaches $k^2$. Finally, we can get a complete category patch $\widetilde{\mathbf{by}}^{s\cup u}$.

4) \textbf{Patch-wise feature generation:} Based on the synthesized category patches, we can generate feature patches for both seen and unseen categories. We use $\widetilde{\mathbf{bw}}^{s\cup u}\in \mathcal{R}^{k\times k\times d}$ to denote the word embedding patch according to an arbitrary category patch $\widetilde{\mathbf{by}}^{s\cup u}$ containing both seen and unseen categories.
As discussed in Sec.~\ref{intro}, we opt for small (\emph{e.g.}, $3\times 3$) patches due to the difficulty in synthesizing reasonable large category patches. Suppose that neighboring pixels within a small patch have similar contextual information, we simply use a same contextual latent code $\mathbf{z} \in \mathcal{R}^{l}$ randomly sampled from $\mathcal{N}(0,1)$ for $k^2$ pixels in the patch. We repeatedly stack $\mathbf{z}$ to construct the contextual latent code patch $\widetilde{\mathbf{bz}} \in \mathcal{R}^{k\times k\times l}$, which is concatenated with $\widetilde{\mathbf{bw}}^{s\cup u}$ as the input of $G$ to generate feature patch $G(\widetilde{\mathbf{bz}},\widetilde{{\mathbf{bw}}}^{s\cup u})$. Note that $G$ consists of several $1\times 1$ conv layers, so each pixel-wise feature in $G(\widetilde{\mathbf{bz}},\widetilde{\mathbf{bw}}^{s\cup u})$ is still generated independently as in Sec.~\ref{sec:context_feature_generator}. Then, synthetic patch-wise features for both seen and unseen categories are used to finetune the classifier (see Sec.~\ref{sec:optimization}).

\subsection{Optimization}\label{sec:optimization}
The procedure of optimizing our CaGNet consists of two steps: 1) training and 2) finetuning, as illustrated in Fig.~\ref{ours}. In the first step, both pixel-wise and patch-wise methods share the same training pipeline on seen data (red line). In the second step, we will introduce pixel-wise finetuning (blue line) and patch-wise finetuning (green line) separately.
In the training step, we utilize training data of only seen categories to update the whole CaGNet $(E, CM, G, D, C)$.
In the finetuning step, we utilize the feature generator $G$ to generate both seen and unseen features, which are used to finetune the classifier $C$, so that CaGNet can generalize to unseen categories. In this step, only $(G,D,C)$ are updated. $(E,CM)$ are frozen since there are no real visual features for gradient backpropagation. Finally, CaGNet only activates $(E,CM,C)$ during the inference, while $(G,D)$ are not used.

\noindent 1) \textbf{Training:}
The objective function of training step can be formulated by 
\begin{eqnarray}\label{eqn:total_loss}
\min\limits_{G,E,C,CM}\max\limits_D \quad \mathcal{L}_{CLS}+\mathcal{L}_{ADV}+\lambda_1 \mathcal{L}_{REC}+\lambda_{2}\mathcal{L}_{KL},
\end{eqnarray}
where all the four loss terms have been discussed in Sec.~\ref{sec:context_feature_generator} and $\lambda_1,\lambda_2$ are hyper-parameters.
The above minimax problem is optimized in an alternating manner.
We first maximize $\mathcal{L}_{ADV}$ to update $ D $ and then minimize the entire objective function to update $\{G,E,C,CM\}$.

\noindent 2) \textbf{Finetuning:}
The finetuning step can be based on either pixel-wise features or patch-wise ones, which are called pixel-wise finetuning and patch-wise finetuning respectively. 

\begin{figure}[tbp]
\centering
\includegraphics[width=\linewidth]{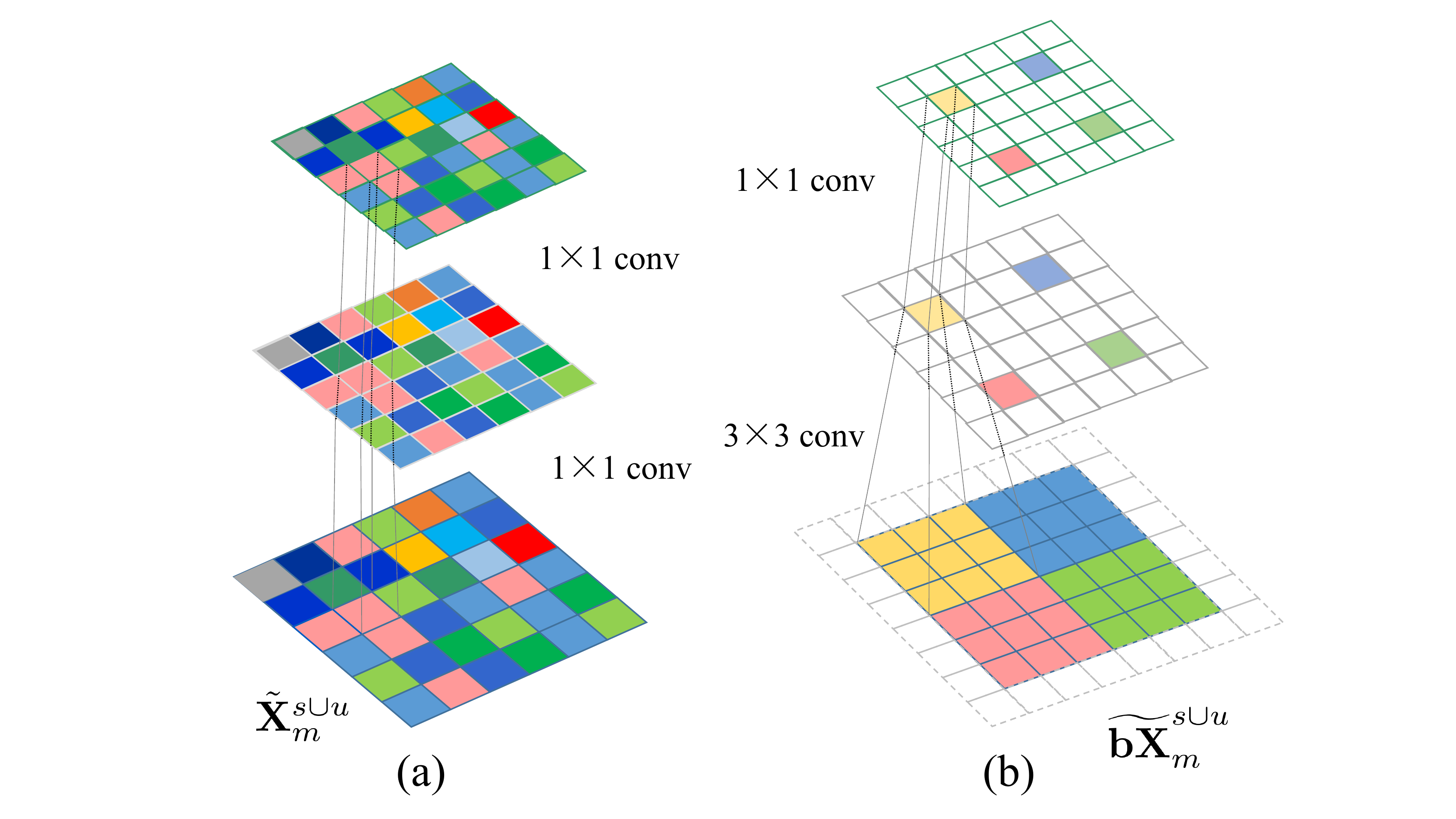}
\caption{The comparison between (a) pixel-wise finetuning and (b) patch-wise finetuning with $k=3$. In (b), we discard the invalid loss entries
when a $3\times3$ conv filter overlaps with more than one generated patch, leading to only four valid loss entries in the loss map.
}
\label{f_losses}
\end{figure}

\textbf{Pixel-wise Finetuning:} For pixel-wise finetuning, the classifier $C$ consists of two $1\times 1$ convolutional (conv) layers as shown in Fig.~\ref{f_losses}, and the discriminator $D$ shares the first $1\times 1$ conv layer with $C$. 
First, we construct the $m$-th synthetic label map ${\tilde{\mathbf{Y}}}^{s\cup u}_m \in \mathcal{R}^{h\times w\times (|\mathcal{C}^s|+|\mathcal{C}^u|)}$ by randomly stacking pixel-wise one-hot label vector ${\tilde{\mathbf{y}}}^{s\cup u}_{m,i}$. As discussed in~\cite{CaGNet}, using approximately the same number of seen pixels and unseen pixels in each label map can generally achieve good performance. The corresponding word embedding map is ${\tilde{\mathbf{W}}}^{s\cup u}_m\in \mathcal{R}^{h\times w\times d}$ with pixel-wise embedding ${\tilde{\mathbf{w}}}^{s\cup u}_{m,i} = {{\mathbf{w}}}_{\tilde{c}_{m,i}^{s\cup u}}$, in which ${\tilde{c}_{m,i}^{s\cup u}}$ is the category label of $i$-th pixel.
Then, we construct latent code map ${\tilde{\mathbf{Z}}}_m$ by randomly stacking pixel-wise latent code ${\tilde{\mathbf{z}}}_{m,i}$, each of which is sampled from $\mathcal{N}(\mathbf{0,1})$ independently. Utilizing ${\tilde{\mathbf{z}}}_{m,i}$ and ${\tilde{\mathbf{w}}}^{s\cup u}_{m,i}$ as the input of generator $G$, we can generate pixel-wise fake features ${\tilde{\mathbf{x}}}^{s\cup u}_{m,i}=G({\tilde{\mathbf{z}}}_{m,i},{\tilde{\mathbf{w}}}^{s\cup u}_{m,i})$, giving rise to fake feature map ${\tilde{\mathbf{X}}}^{s\cup u}_m$ as shown in Fig.~\ref{f_losses}.
The objective function of pixel-wise finetuning can be written as
\begin{eqnarray}\label{opt:finetune} \min\limits_{G,C}\max\limits_D \quad \mathcal{\tilde{L}}_{CLS}+\mathcal{\tilde{L}}_{ADV},\end{eqnarray}
in which classification loss $\mathcal{\tilde{L}}_{CLS}$ and adversarial loss $\mathcal{\tilde{L}}_{ADV}$ are obtained by modifying (\ref{eqn:L_cls}) and (\ref{eqn:L_adv}) respectively:
\begin{eqnarray}\label{eqn:L_cls_finetune}
\mathcal{\tilde{L}}_{CLS} =-\sum_{m,i}{\tilde{\mathbf{y}}}_{m,i}^{s\cup u}\log(C({\tilde{\mathbf{x}}}_{m,i}^{s\cup u})),
\end{eqnarray}
\begin{eqnarray}\label{eqn:L_adv_finetune}
\mathcal{\tilde{L}}_{ADV} =\sum_{m,i} (1-D(G({\tilde{\mathbf{z}}}_{m,i},{\tilde{\mathbf{w}}}^{s\cup u}_{m,i})))^2.
\end{eqnarray}

\textbf{Patch-wise Finetuning:} As mentioned in Sec.~\ref{sec:pixelcnn++}, we generate small category patches, resulting in small feature patches. In our implementation, we synthesize $3\times3$ feature patches by default, so we will take $k=3$ as an example to describe patch-wise finetuning. In this case, the classifier $C$ consists of a $3\times 3$ conv layer ($stride$=$1$) and a $1\times 1$ conv layer as shown in Fig.~\ref{f_losses}. The discriminator $D$ shares the first $3\times 3$ conv layer with $C$. 

One issue is that most of generated category patches in Sec.~\ref{sec:pixelcnn++} contain more than one category. However, in reality, the majority of $3\times 3$ patches in the training data only have a single category.
For ease of description, we name the category patches with only one category as pure patches and the generated patches with multiple categories as mixed patches. We fuse pure patches with mixed patches at the ratio of $\lambda_{r}:1$ to better imitate real training data, where $\lambda_{r}\in\mathcal{R}$ is a hyper-parameter.

It is noteworthy that we encounter two main problems when dealing with mixed patches. One problem is that real category patches in the training data usually contain no more than $3$ categories, but the generated mixed patches often contain more than $3$ categories.
Another problem is that the generated mixed patches contain very few unseen pixels due to seen bias. To tackle the above two problems, we use a simple yet effective strategy to control the number of categories and simultaneously ensure more unseen pixels in mixed patches. To be specific, we preset the first $\lambda_{f}$ pixels in each category patch as a same random unseen category $\widetilde{c}^u\in \mathcal{C}^u$, where $\lambda_{f}\in\{0,1,2,\cdots,k^2-1\}$ is a hyper-parameter.
Then, PixelCNN will sequentially generate the remaining pixels $\{\lambda_{f}+1,\lambda_{f}+2,\cdots,k^2\}$ based on $\lambda_{f}$ observed pixels, which is analogous to category map inpainting.
Finally, we filter out the mixed patches with more than $3$ categories.

The procedure of patch-wise finetuning is similar to that of pixel-wise finetuning. First, we construct the $m$-th label map $\tilde{\mathbf{Y}}_{m}^{s\cup u}$ by randomly stacking generated category patches $\widetilde{\mathbf{by}}_{m,i}^{s\cup u}$. For each category patch $\widetilde{\mathbf{by}}_{m,i}^{s\cup u}$, we can obtain the corresponding word embedding patch $\widetilde{\mathbf{bw}}_{m,i}^{s\cup u}$. The associated latent code patch ${\widetilde{\mathbf{bz}}}_{m,i}$ is obtained by repeatedly stacking a same random latent code $\mathbf{z}$.  
Then, we construct the $m$-th word embedding map $\tilde{\mathbf{W}}_{m}^{s\cup u}$ (\emph{resp.}, latent code map ${\tilde{\mathbf{Z}}}_{m}$) by randomly stacking $\widetilde{\mathbf{bw}}_{m,i}^{s\cup u}$ (\emph{resp.}, ${\widetilde{\mathbf{bz}}}_{m,i}$).
Based on $\tilde{\mathbf{W}}_{m}^{s\cup u}$ and  ${\tilde{\mathbf{Z}}}_{m}$, we can generate fake feature map ${\tilde{\mathbf{X}}}_{m}^{s\cup u}$ with feature patches ${\widetilde{\mathbf{bx}}}_{m,i}^{s\cup u}$. Recall that the feature generator $G$ consists of several $1\times 1$ conv layers, so each feature patch is actually generated independently, that is, ${\widetilde{\mathbf{bx}}}^{s\cup u}_{m,i}=G({\widetilde{\mathbf{bz}}}_{m,i},{\widetilde{\mathbf{bw}}}^{s\cup u}_{m,i})$.

The objective function of patch-wise finetuning has the same form as (\ref{opt:finetune}) with slightly different notations. In detail, we replace ${\tilde{\mathbf{y}}}_{m,i}^{s\cup u}$ (\emph{resp.}, ${\tilde{\mathbf{x}}}_{m,i}^{s\cup u}$) in (\ref{eqn:L_cls_finetune}) with ${\widetilde{\mathbf{by}}}_{m,i}^{s\cup u}$(\emph{resp.}, ${\widetilde{\mathbf{bx}}}_{m,i}^{s\cup u}$), 
and replace ${\tilde{\mathbf{w}}}_{m,i}^{s\cup u}$ (\emph{resp.}, ${\tilde{\mathbf{z}}}_{m,i}$) in (\ref{eqn:L_adv_finetune}) with ${\widetilde{\mathbf{bw}}}_{m,i}^{s\cup u}$(\emph{resp.}, ${\widetilde{\mathbf{bz}}}_{m,i}$). The output adversarial loss map and classification loss map are of the same size as the input feature map ${\tilde{\mathbf{X}}}_{m}^{s\cup u}$.
Note that in the $3\times 3$ conv layer ($stride$=$1$), when a conv kernel overlap with more than one generated patch, the covered $3\times 3$ patch may be an unreasonable patch, leading to an invalid loss entry in the loss map. 
Therefore, we apply a binary loss mask to the output loss
map to discard the invalid loss entries.
For example, as shown in Fig.~\ref{f_losses}, a $6\times 6$ feature map results in a $6\times 6$ loss map, but only $4$ loss entries are valid.

\noindent\textbf{Optimization strategy:} We use ResNet-101~\cite{resnet-101} pre-trained on ImageNet~\cite{ILSVRC15} to initialize Deeplabv2 backbone $E$. Then, the training step is applied until our model converges. After that, the training step and the finetuning step are performed alternatingly every 100 iterations so that the segmentation model can be gradually adapted to unseen categories. If we perform finetuning continuously, the model will be biased towards fake features and the final segmentation quality will be impaired. 



\section{Experiments}

\begin{table*}
\centering
\setlength{\tabcolsep}{1mm}{
\begin{tabular}{c|cccc|ccc|ccc}
\toprule[1.5pt]
\multicolumn{11}{c}{Pascal-VOC}\\ \hline
\multirow{2}*{Method} & \multicolumn{4}{c}{Overall}   & \multicolumn{3}{c}{Seen}    & \multicolumn{3}{c}{Unseen}      \\ 
~     & \textbf{hIoU} & mIoU &  pixel acc.& mean acc. & mIoU  & pixel acc. &  mean acc. & mIoU  & pixel acc. &  mean acc. \\ \hline
SPNet~\cite{xian2019semantic}& 0.0002 &0.5687 & 0.7685  & 0.7093 &0.7583  & \textbf{0.9482}    & \textbf{0.9458} & 0.0001 &   0.0007  &   0.0001  \\ 
SPNet-c~\cite{xian2019semantic}&0.2610  &0.6315  & 0.7755  &0.7188  &0.7800  &0.8877     &0.8791  & 0.1563 &0.2955 &0.2387 \\ 
ZS3Net~\cite{bucher2019zero}&0.2874  &0.6164  & 0.8107  &0.7349  &0.7730  & 0.9296    &0.8772  &0.1765  &0.3047     &0.2468     \\ 
CSRL~\cite{li2020consistent}& 0.3712 & 0.6458 & \textbf{0.8132}  &0.7512  &0.7809  & 0.9126    & 0.8845 &0.2435  & 0.3876    & 0.3431    \\ 
Hu \emph{et al.}~\cite{hu2020uncertainty}& 0.2682 &0.6210  & 0.7811  &0.7208  & 0.7771 &0.8993     & 0.8659 &0.1621  & 0.2759    & 0.2276    \\
CaGNet(pi)~\cite{CaGNet}&0.3972  &0.6545   & 0.8068   &0.7636   & \textbf{0.7840}  & 0.8950     & 0.8868  &0.2659   & 0.4297     &  0.3940    \\
\textbf{CaGNet(pa)}&\textbf{0.4326}  &\textbf{0.6623}  & 0.8068  &\textbf{0.7643} & 0.7814 &  0.8745   &0.8621& \textbf{0.2990} & \textbf{0.5176} & \textbf{0.4710}         \\ \hline
ZS3Net+ST~\cite{bucher2019zero}& 0.3328 & 0.6302 & 0.8095  &  0.7382&0.7802 &\textbf{0.9189}  &\textbf{0.8569}       & 0.2115 &  0.3407  & 0.2637    \\
CaGNet(pi)+ST~\cite{CaGNet} &0.4366  &0.6577   & \textbf{0.8164}  &0.7560  &\textbf{0.7859}   & 0.8704    &0.8390  & 0.3031  &  0.5855    &   0.5071     \\
\textbf{CaGNet(pa)+ST} &\textbf{0.4528}  &\textbf{0.6657}   & 0.8036  &\textbf{0.7650}  &0.7813   & 0.8527    &0.8394  & \textbf{0.3188}  &  \textbf{0.5939}    &   \textbf{0.5417}     \\ \toprule[1.2pt]
\multicolumn{11}{c}{COCO-stuff}\\ \hline
SPNet~\cite{xian2019semantic}& 0.0140 & 0.3164 & 0.5132  &  0.4593&0.3461  &   \textbf{0.6564}  & 0.5030 &0.0070  & 0.0171    &  0.0007   \\ 
SPNet-c~\cite{xian2019semantic}&0.1398  & 0.3278 &0.5341   &0.4363  & 0.3518 &  0.6176   &0.4628  & 0.0873 &0.2450     & 0.1614    \\ 
ZS3Net~\cite{bucher2019zero}& 0.1495 & 0.3328 & 0.5467  &0.4837  &0.3466  & 0.6434    & 0.5037 & 0.0953 & 0.2275    &   0.2701  \\ 
CSRL~\cite{li2020consistent}& 0.1657 & 0.3312 & 0.5583  & 0.4843 & 0.3468 &0.6502     &0.5078  &0.1089  &0.2331    & 0.2736    \\ 
Hu \emph{et al.}~\cite{hu2020uncertainty}& 0.1101 & 0.3212 & 0.5532  &0.4475  &0.3501  & 0.6532    & 0.5050 & 0.0653 & 0.1954    & 0.1508    \\
CaGNet(pi)~\cite{CaGNet}&  0.1819 &\textbf{0.3345}  & \textbf{0.5658} &0.4845  & \textbf{0.3549} &  0.6562   & 0.5066 & 0.1223 & 0.2545    &  0.2701   \\
\textbf{CaGNet(pa)}&\textbf{0.1984}  &0.3327  & 0.5632  &\textbf{0.4909} & 0.3468&  0.6542   & \textbf{0.5125} & \textbf{0.1389} & \textbf{0.2962}    &\textbf{0.3132}     \\ \hline
ZS3Net+ST~\cite{bucher2019zero}& 0.1620 &0.3367  & 0.5631  & \textbf{0.4862} &0.3489  & 0.6584    & 0.5042 & 0.1055  & 0.2488  &0.2718     \\
CaGNet(pi)+ST~\cite{CaGNet} &0.1946  &0.3372   & 0.5676  &0.4854  &0.3555   & \textbf{0.6587}    &\textbf{0.5058}  & 0.1340  &  0.2670    &   \textbf{0.2728}     \\
\textbf{CaGNet(pa)+ST} &\textbf{0.2269}  &\textbf{0.3456}   & \textbf{0.5944}    &0.4629   & \textbf{0.3617}    &0.6391  & 0.4828  &  \textbf{0.1654}    &   \textbf{0.4404}  &  0.2567  \\ \toprule[1.2pt]
\multicolumn{11}{c}{Pascal-Context}\\ \hline
SPNet~\cite{xian2019semantic}& 0& 0.2938& 0.5793& 0.4486& 0.3357& \textbf{0.6389}& 0.5105& 0&0& 0\\ 
SPNet-c~\cite{xian2019semantic}& 0.0718&0.3079& 0.5790& 0.4488& 0.3514&  0.6213& 0.4915&  0.0400& 0.1673&  0.1361\\ 
ZS3Net~\cite{bucher2019zero}&0.1246  & 0.3010 &0.5710   & 0.4442 & 0.3304  &   0.6099  & 0.4843 & 0.0768 & 0.1922    &  0.1532   \\ 
CSRL~\cite{li2020consistent}& 0.1970 & 0.3308 &  0.5869 &0.4863  & 0.3575 & 0.6175    & 0.5079 &0.1358  & 0.2893    & 0.3156    \\ 
Hu \emph{et al.}~\cite{hu2020uncertainty}& 0.1056 & 0.3159 & 0.5662  &0.4325  &0.3346  & 0.6071    &0.4922  & 0.0627 & 0.1683    &0.1433     \\
CaGNet(pi)~\cite{CaGNet}&0.2061  &\textbf{0.3347}  & \textbf{0.5975}  &0.4900 & 0.3610 & 0.6180    & 0.5140 & 0.1442 & 0.3976    &0.3248     \\
\textbf{CaGNet(pa)}&\textbf{0.2135}  &0.3243  & 0.5816  &\textbf{0.5082} & \textbf{0.3718} &  0.6004   & \textbf{0.5282} & \textbf{0.1498} & \textbf{0.3981}    &\textbf{0.3412}     \\ \hline
ZS3Net+ST~\cite{bucher2019zero}& 0.1488 & 0.3102 & 0.5842  & 0.4532 & 0.3398 & 0.6107    &  0.4935&0.0953  &  0.3030  &   0.1721  \\
CaGNet(pi)+ST~\cite{CaGNet} &0.2252  &0.3352   & \textbf{0.5951}  &0.4962  &\textbf{0.3644}   & \textbf{0.6120}    &0.5065  & 0.1630  &  0.4038    &   \textbf{0.4214} \\
\textbf{CaGNet(pa)+ST} &\textbf{0.2478}  &\textbf{0.3364}   & 0.5832  & \textbf{0.4964} &0.3482   & 0.6010    &\textbf{0.5119}  & \textbf{0.1923}  &  \textbf{0.4075}    &   {0.4023}     \\ \bottomrule[1.5pt]
\end{tabular}
\caption{Zero-shot segmentation performances on Pascal-VOC, COCO-stuff, and Pascal-Context. ``pi" (\emph{resp.}, ``pa'') means pixel-wise (\emph{resp.}, patch-wise) finetuning. ``ST'' stands for self-training. The best results with or w/o self-training are denoted in boldface, respectively.}
\label{coco}
}
\end{table*}

\subsection{Datasets and Basic Settings}

Our experiments are conducted on Pascal-VOC 2012~\cite{EveringhamThePascalVisual}, Pascal-Context \cite{mottaghi2014role}, and COCO-stuff~\cite{caesar2018coco} dataset. 1) Pascal-VOC 2012 contains 1464 training images with segmentation annotations of 20 object categories. Following \cite{xian2019semantic,bucher2019zero}, we adopt additional supervision from semantic boundary annotations~\cite{Hariharan2011Semantic} for Pascal-VOC. 2) The Pascal-Context dataset contains 4998 training images and 5105 validation images of 33 object/stuff categories. 3) COCO-stuff has 164K images with dense pixel-wise annotations from 182 categories. We follow the standard train/test set split of three datasets. The seen/unseen category split can be found in the conference version~\cite{CaGNet}. In the training stage, we only use the pixel-wise annotations of seen categories and ignore other pixels. 

In terms of word embedding,  we concatenate $300$-dim word2vec~\cite{mikolov2013distributed} (trained on Google News) and  $300$-dim fast-Text~\cite{joulin2017bag} (trained on Common Crawl), leading to $600$-dim word embedding. If a category name has multiple words, we average the embeddings of multiple words to obtain the word embedding of this category. 

The experimental results for seen categories or unseen categories are analysed based on four evaluation metrics: pixel accuracy, mean accuracy, mean IoU (mIoU), and harmonic IoU (hIoU) following \cite{xian2019semantic,CaGNet}.
Among these evaluation metrics, ``mIoU'' quantizes the overlap between predicted objects and ground-truth ones, which is more reliable than ``accuracy'' considering the integrity of objects. In zero-shot segmentation, ``hIoU'' is more valuable than ``mIoU'' because it can balance all categories and prevent seen categories from dominating the overall results.

\subsection{Implementation Details}\label{hypers}

We use Deeplabv2~\cite{chen2018deeplab} with ResNet-101 as segmentation backbone $E$. 
The generator $ G $ has three $1\times 1$ convolutional (conv) layers ($512$ intermediate dimension, Leaky ReLU, and dropout for each layer). For our pixel-wise based method, Classifier $ C $ and discriminator $ D $ have a shared $1\times 1$ conv layer~\cite{xu2018pad-net:} and a seperate $1\times 1$ conv layer. For our patch-wise based method with patch size $k\times k$ ($k=3$ by default), we change the conv layer shared by $C$ and $D$ to $[\frac{k-1}{2}]$ conv layers ($kernel\, size=3$, $stride=1$,  $padding=1$). 
For PixelCNN,  we use $4$ masked conv layers (1A+3B).

In the training stage, the learning rate is initialized as $2.5e^{-4}$ and divided by $10$ whenever the loss stops decreasing. Most experiments have been carried out with batch size $8$ on one Tesla V100. The side length of input images is set to $368$. 
We set $\lambda_{1}=10,\lambda_{2}=100$ in (\ref{eqn:total_loss}) based on the validation set made by splitting out 20\% seen categories as validation categories~\cite{CaGNet}.
Specifically, we split out 20\% seen categories as the validation categories. Then we separate the training set into sub-training set (80\%) and validation set (20\%). Note that in the sub-training set, the pixel-level annotations for validation categories are ignored. We train our model based on sub-training set and the semantic representations of seen categories. Then, the trained model is applied to the validation set to determine the optimal hyper-parameters based on the validation performance.
For category patch generation, we set $\lambda_{f}=5,\lambda_{r}=4$, which will be further analysed in Sec.~\ref{Hyper-parameter_Analyses}.

\subsection{Comparison with State-of-the-art}\label{Implement}

We refer to our pixel-wise based method and patch-wise based method as CaGNet(pi) and CaGNet(pa), respectively.
We compare with four baselines: SPNet~\cite{xian2019semantic},  ZS3Net~\cite{bucher2019zero}, CSRL~\cite{li2020consistent}, and Hu \emph{et al.}~\cite{hu2020uncertainty}. We also compare with SPNet-c, \emph{i.e.}, an extension of SPNet\footnote{Our reproduced results of SPNet on Pascal-VOC dataset are obtained using their released model and code with careful tuning, but still lower than their reported results.}, which deducts the prediction scores of seen categories by a calibration factor. ZS3Net(GC) in~\cite{bucher2019zero} is not included for comparison, because the object relational graphs they used are unavailable in our setting and difficult to acquire in real-world applications. 
In addition, we employ the Self-Training (ST) strategy in~\cite{bucher2019zero} for both ZS3Net and our CaGNet. To be exact, we use the trained segmentation model to tag unlabeled pixels in training images and then use the new training set to finetune our CaGNet. The above two steps are executed iteratively. Note that previous works~\cite{xian2019semantic,CaGNet,bucher2019zero,li2020consistent,hu2020uncertainty} reported results in considerably different settings (\emph{e.g.}, backbone, word embedding, category split). We mainly follow the setting in~\cite{xian2019semantic,CaGNet} and compare all methods in the same setting for fairness. We also compared our CaGNet(pi) with ZS3Net~\cite{bucher2019zero} in its setting in our conference version~\cite{CaGNet}, so we omit the results here. 

Table~\ref{coco} summarizes the experiment results of all methods, which shows that our method achieves significant improvements for ``unseen'' and ``overall'' evaluation metrics, especially \emph{w.r.t.} ``mIoU'' and ``hIoU''. Although our method underperforms SPNet in some ``seen'' evaluation cases, our method sacrifices some seen pixels for much better overall performance. 
We also observe that our CaGNet(pa) outperforms CaGNet(pi) \emph{w.r.t.} ``overall" hIoU and ``Unseen" mIoU (two most valuable metrics) on all three datasets, especially on the datasets with fewer categories like Pascal-VOC. This demonstrates the advantage of patch-wise finetuning, which takes inter-pixel relationship into consideration.

\begin{table}[!tbp]
\centering
\begin{tabular}{c|cccc}
\toprule[1.5pt]
 &\textbf{hIoU} & mIoU& S-mIoU& U-mIoU    \\ \hline \hline
pure   &0.4104 &0.6421 &0.7715&  0.2795    \\
mixed (random)& 0.4027&0.6397 &0.7801&  0.2713      \\
mixed (constrained random)&0.4205 &0.6553 &0.7743&0.2886 \\
mixed (PixelCNN)&0.4163 & 0.6542&0.7812&0.2837\\
mixed (constrained PixelCNN)&0.4268 &0.6601 &0.7756&0.2943 \\
\textbf{Ours: pure + mixed}  & \textbf{0.4326} & \textbf{0.6623} & \textbf{0.7833}&\textbf{0.2988}   \\\bottomrule[1.5pt]
\end{tabular}
\caption{Ablation studies of special cases of patch-wise finetuning on Pascal-VOC.}
\label{bfablation}
\end{table}
%
%
%
%
%
%
\begin{figure}[tbp]
\centering
\includegraphics[width=\linewidth]{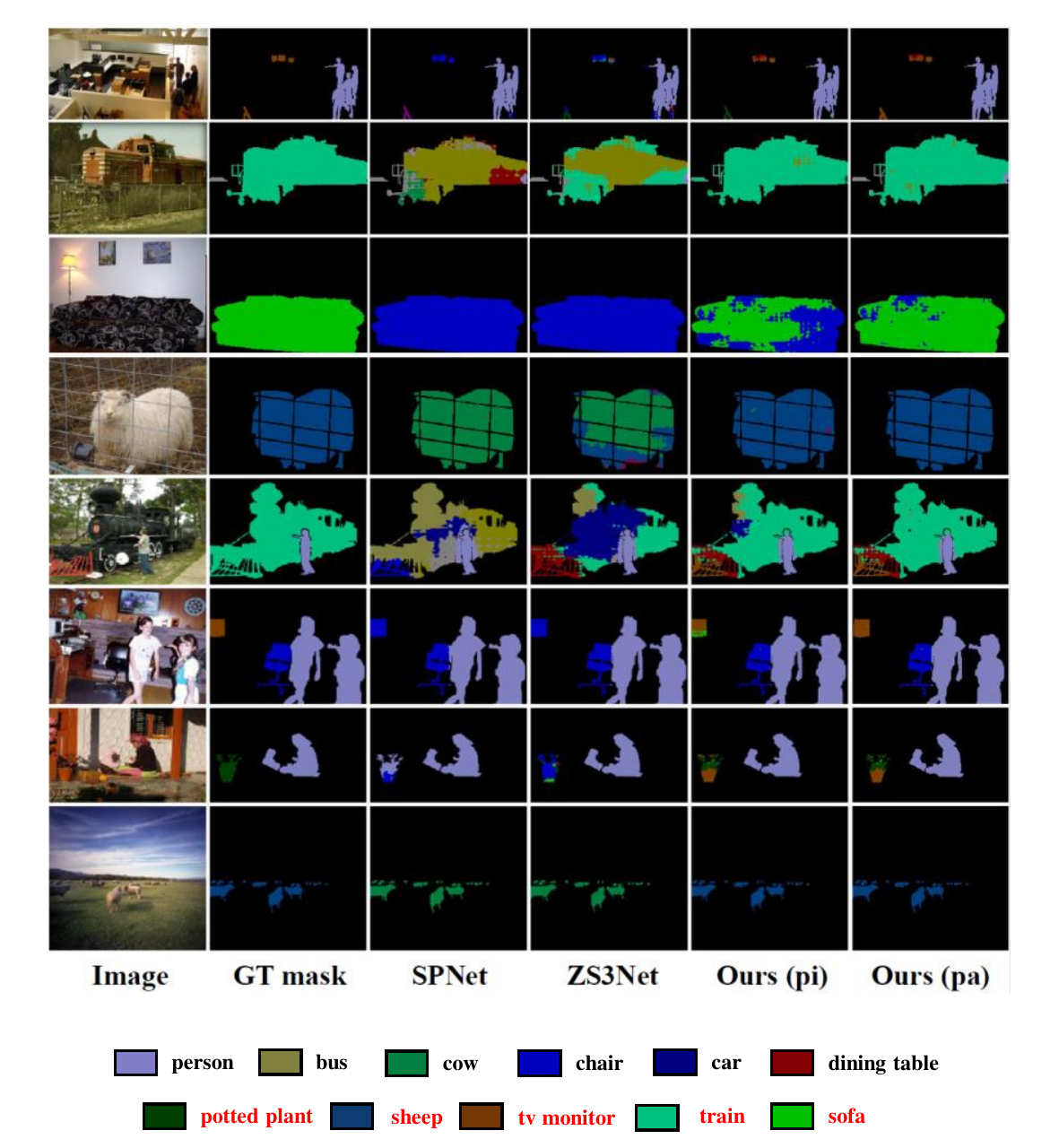}
\caption{Visualization of zero-shot segmentation results on Pascal-VOC. GT mask is ground-truth segmentation mask. The categories are marked at the bottom, where red color denotes unseen ones.}
\label{visseg}
\end{figure}

\subsection{Ablation Studies}\label{sec:blocka}
The results of ablation studies for our CaGNet(pi) have been reported in our c'
onference version~\cite{CaGNet}, so we omit them here due to page limitation.

To verify the performance gain of patch-wise finetuning in our CaGNet(pa), we conduct ablation studies on Pascal-VOC.
Recall that we use two types of category patches, \emph{i.e.}, pure patches and mixed patches, to synthesize feature patches. Pure patches only contain a single category, while mixed patches are generated by our PixelCNN and contain multiple categories. Instead of PixelCNN, we can also randomly construct mixed patches with multiple categories. In Table~\ref{bfablation}, we compare the  segmentation results when finetuning the classifier with pure, mixed (random), or mixed (pixelCNN) patches. As mentioned in Sec.~\ref{sec:optimization}, we impose two constraints on the mixed patches: 1) preset the first $\lambda_{f}$ pixels; 2) remove the mixed patches with more than $3$ categories. We also compare the mixed patches with or without constraints in Table~\ref{bfablation}. 

The first row obtains slight improvement over our pixel-wise based method in Table~\ref{coco}, which shows that patch-wise finetuning is beneficial even with pure patches. 
By comparing the third (\emph{resp.}, fifth) with the second (\emph{resp.}, fourth) row, we observe that adding constraints can significantly improve the performance, because the abovementioned two constraints can improve the authenticity of category patches and the generalization performance on unseen categories. The fifth row achieves the best results among mixed patches and also significantly outperforms pure patches, which demonstrates the advantage of mixed patches generated by PixelCNN under necessary constraints. The last row is our method, which fuses pure patches with mixed patches (constrained PixelCNN) at the ratio of $\lambda_r:1$ for patch-wise finetuning. This method achieves the best results because both pure and mixed patches commonly exist in the real training data.


\subsection{Hyper-parameter Analyses}\label{Hyper-parameter_Analyses}



We have analysed $\lambda_{1}$ and $\lambda_{2}$ in (\ref{eqn:total_loss}) in the conference version~\cite{CaGNet}, so we omit them here due to page limitation.

\begin{figure}[tbp]
\centering
\includegraphics[width=\linewidth]{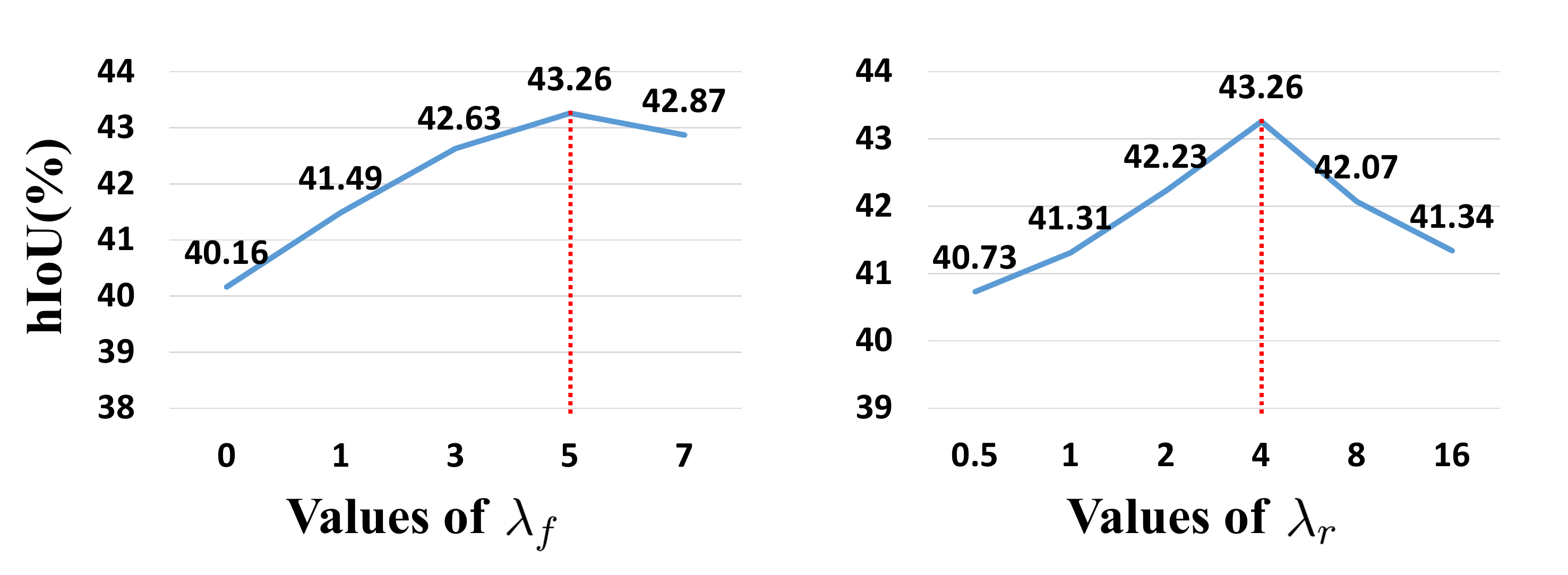}
\caption{The performance variance of CaGNet(pa) when varying $\lambda_f,\lambda_r$ on Pascal-VOC. The dashed lines denote the default values used in our paper. }\label{alpha4}
\end{figure}

In patch-wise finetuning, we explore the impact of two hyper-parameters $\lambda_f,\lambda_r$. We vary $\lambda_f$ (\emph{resp.}, $\lambda_{r}$) within the range [0,8] (\emph{resp.}, [0.5,16]) and report hIoU(\%) in Fig.~\ref{alpha4}. 
The hIoU increases when $\lambda_f$ increases from $0$ to $5$, which indicates the benefit of introducing more unseen information. When $\lambda_f$ is larger than $5$, the performance decreases due to the lack of patch variety, because preset pixels dominate the patch and only leave few pixels for generation. When $\lambda_r$ is too large, pure patches account for a large proportion, in which case inter-pixel relationship across different categories is seldom considered. When $\lambda_r$ is too small, mixed patches will overwhelm pure patches, which contradicts the fact that most real category patches in the training data are pure patches. Consequently, it harms the performance on both seen and unseen categories. Based on Fig.~\ref{alpha4}, the optimal choice is $\lambda_f=5$ and $\lambda_r=4$. We also observe that this choice can consistently achieve the best results on all three datasets.

\begin{table}[t]
\centering
\begin{tabular}{c|ccc}
\toprule[1.5pt]
patch size &Pascal-VOC & COCO-stuff& Pascal-Context    \\ \hline \hline
 $k=3$   &  \textbf{0.4326} &  \textbf{0.1984} &   \textbf{0.2135}  \\
  $k=5$  &0.4065 & 0.1851 & 0.2071   \\
  $k=7$  & 0.3767  & 0.1575& 0.1772  \\\bottomrule[1.5pt]
\end{tabular}
\caption{Performances (\textbf{hIoU}) of CaGNet(pa) with different patch size $k$ on three datasets.}
\label{blocksize}
\end{table}

\subsection{Different Patch Sizes}
As discussed in Sec.~\ref{sec:pixelcnn++}, we opt for small patches and set $k=3$ by default.
On all three datasets, we have consistent observation that the performance decreases as the patch size $k$ increases.
The unsatisfactory performance with large $k$ might be caused by the following two reasons: 1) Generating plausible large-scale category patches is very challenging, because the shapes and poses of unseen objects are very hard to imagine; 2) The contextual information of pixels within a large patch will be different from each other, so using a same contextual latent code for all pixels in a patch becomes unreasonable. 

\begin{figure}[tbp]
\centering
\includegraphics[width=\linewidth]{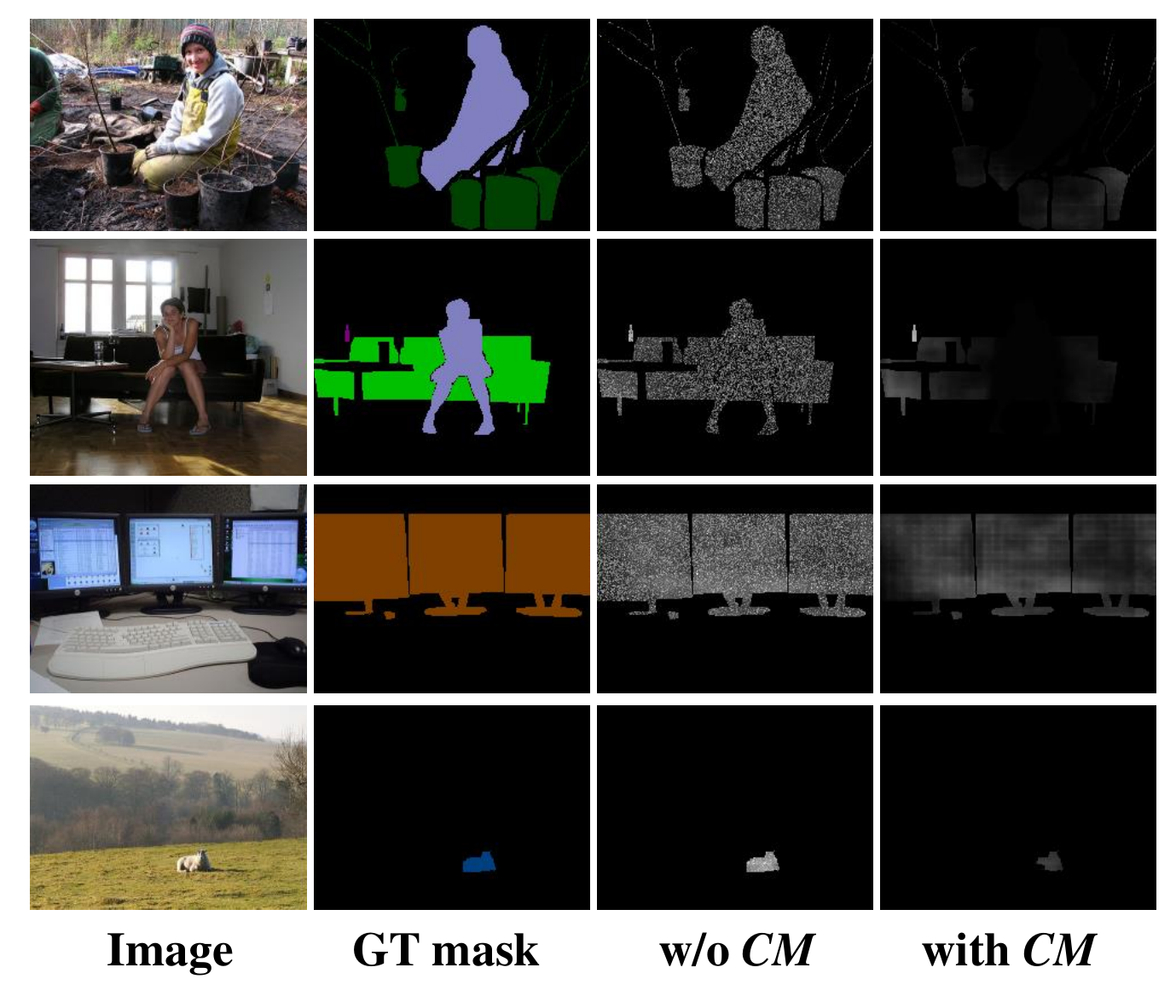}
\caption{Visualization of feature generation quality of CaGNet(pa) on Pascal-VOC test set. GT mask is ground-truth segmentation mask. In the third and fourth columns, we show the reconstruction loss maps between  generated feature maps and real feature maps (the darker, the better).}
\label{visgen}
\end{figure}

\subsection{Computation Cost and Processing Time}
We report the inference FLOP and FPS of different methods in Table~\ref{speed}. We run all methods on pytorch with a single NVIDIA V100. The input image size is $368\times 368$.  Note that during inference, only backbone $E$, contextual module $CM$, and classifier $C$ are used. From the table, we can observe that SPNet holds the smallest FLOP while the FLOPs of other methods increase as the model complexity increases.

Compared with ZS3Net, our CaGNet has an extra contextual module. Thus it takes a little bit more time to feed-forward an image than ZS3Net. The difference between CaGNet(pi) and CaGNet(pa) is only the kernel size (1 \emph{vs.} 3) of classifier. As a result, CaGNet(pi) and CaGNet(pa) are very close in terms of efficiency.

\begin{table}[t]
\centering
\begin{tabular}{c|cc}
\toprule[1.5pt]
Method &FLOP & FPS  \\ \hline \hline
SPNet~\cite{xian2019semantic} & 99.04G & 15.8  \\
ZS3Net~\cite{bucher2019zero} & 102.71G& 11.6  \\
CSRL~\cite{li2020consistent} & 101.58G & 12.1  \\
Hu \emph{et al.}~\cite{hu2020uncertainty} &99.65G & 12.7 \\
CaGNet(pi) &102.98G  &10.5  \\
CaGNet(pa) & 103.25G &10.2  \\\bottomrule[1.5pt]
\end{tabular}
\caption{FLOP and FPS of different methods.}
\label{speed}
\end{table}

\subsection{Diversity of Generated Unseen Features}
As claimed in Section~\ref{intro}, we attempt to mitigate the mode collapse problem by establishing the one-to-one correspondence between pixel-wise contextual information and pixel-wise feature. 
So we use contextual latent code to replace random vector and employ feature reconstruction loss. To verify this hypothesis, we compare the diversity of  unseen features generated by CaGNet(pi) and ZS3Net on Pascal-VOC. We also compare the diversity with real unseen features. For a fair comparison, we sample 1000 random vectors from $\mathcal{N}(\mathbf{0,1})$, which are fed into the feature generator of ZS3Net and CaGNet to generate 1000 fake features for each unseen category. We calculate the averaged pairwise Euclidean distance among the features within each unseen category, and then calculate the average over all unseen categories. We also sample 1000 real features for each unseen category and calculate the averaged pairwise distance in the same way. 

The results are summarized in Table~\ref{diversity}. We can observe that the diversity of real unseen features in CaGNet(pi) and ZS3Net are similar. For the generated unseen features, the diversity of CaGNet(pi) is much larger than ZS3Net and closer to real features, which demonstrates that our CaGNet can generate more diverse unseen features and thus improve the segmentation performance on unseen categories. 

\begin{table}[t]
\centering
\setlength{\tabcolsep}{1mm}{
\begin{tabular}{c|c|ccccc|c}
\toprule[1.5pt]
Method& Feature  & plant& sheep & monitor& train& sofa & Average  \\ \hline 
ZS3Net& fake& 810.5 &850.2 &	623.1 &	1000.3 &	710.2 & 	798.9 \\
CaGNet& fake&1280.9 &	1254.2	 &1100.5 &	1461.4	&1175.3 & 	1254.5 \\ \hline 
ZS3Net& real&2674.4 &	1940.1	&2254.3	   & 2514.3	&2671.4	 & 2410.9 \\
CaGNet& real& 2678.1&	1937.1	 & 2258.1	& 2519.7& 	2665.0	&2411.6   \\\bottomrule[1.5pt]
\end{tabular}
\caption{Average pairwise distance among real features and generated fake features for unseen categories on Pascal-VOC.}
\label{diversity}
}
\end{table}

\subsection{Qualitative Analyses}

In this section, we provide some visualization results of our method on Pascal-VOC.

\noindent\textbf{Semantic segmentation:} Fig.~\ref{visseg} shows some segmentation results of baselines and our method. ``GT'' means ground-truth segmentation mask. Our CaGNet(pi) and CaGNet(pa) perform favorably on unseen objects, \emph{e.g.}, tv monitor (orange), train (light green), sofa (green), potted plant (dark green), sheep (dark blue), as shown at the bottom of Fig.~\ref{visseg}. 
Moreover, our CaGNet(pa) can segment unseen objects better than CaGNet(pi), \emph{e.g.}, sofa in the third row and tv monitor in the sixth row, showing that patch-wise finetuning is helpful.


\noindent \textbf{Feature generation:} We evaluate the effectiveness of feature generation of CaGNet(pa)  on test images. On the one hand, one test image is fed into the segmentation backbone to obtain the real feature map. On the other hand, the corresponding word embedding and the latent code are fed into the generator to obtain the generated feature map. Then, we calculate the reconstruction loss map based on these two feature maps, as shown in Fig.~\ref{visgen}, where smaller loss (darker pixel) implies better generation quality.
``With $CM$'' represents our full method (latent code is contextual latent code produced by $CM$) and ``w/o $CM$'' is a controlled experiment (latent code is random vector). We can see that our $CM$ is able to generate better features for seen categories (\emph{e.g.}, ``person'') and unseen categories (\emph{e.g.}, ``potted plant'', ``sofa'', ``tv monitor'', ``sheep'') consistently.


\begin{figure}[tbp]
\centering
\includegraphics[width=0.9\linewidth]{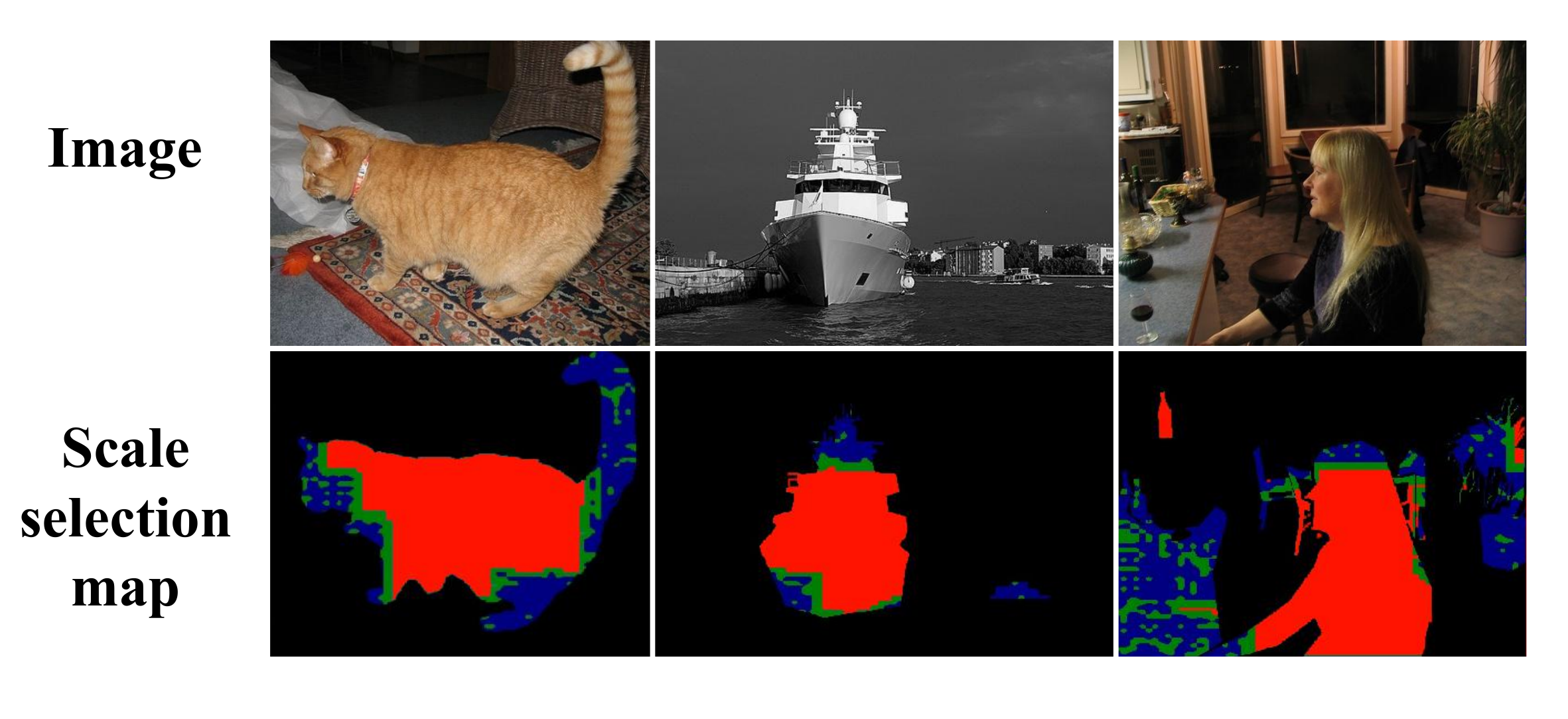}
\caption{Visualization of scale selection in CaGNet(pi) on Pascal-VOC. 
GT mask is the ground-truth segmentation mask.
In scale selection maps, blue, green, red means selecting small-scale, middle-scale, and large-scale context respectively.}
\label{viscon}
\end{figure}

\begin{figure*}[tbp]
\centering
\includegraphics[width=\linewidth]{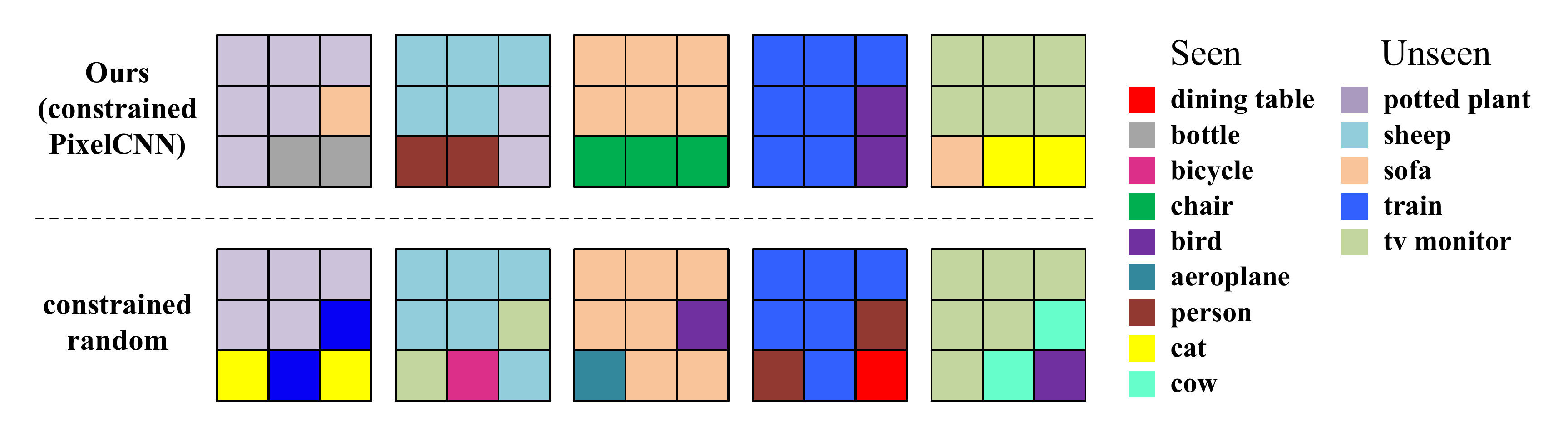}
\caption{Visualization of mixed patches (constrained PixelCNN) and mixed patches (constrained random) on Pascal-VOC when $k=3$ and $\lambda_f = 5$. }
\label{reasonable}
\end{figure*}

\noindent\textbf{Context selector:} A context selector is proposed in our contextual module to select the context of suitable scale for each pixel. We calculate the scale weight map $[\textbf{A}_n^0,\textbf{A}_n^1,\textbf{A}_n^2]\in\mathcal{R}^{h\times w\times 3}$ which contains three scale weights for each pixel in the feature map $\mathbf{F}_n$.
$\mathbf{A}_n^0$ (\emph{resp.}, $\mathbf{A}_n^1$, $\mathbf{A}_n^2$) represents small scale (\emph{resp.}, middle scale, large scale) with $3\times3$ (\emph{resp.}, $7\times7$, $17\times17$) receptive field size \emph{w.r.t.} $\mathbf{F}_n$, according to the calculation method in \cite{luo2016understanding}. 
By choosing the scale with the largest weight for each pixel, a scale selection map can be obtained.

As shown in Fig.~\ref{viscon}, we present some images with their scale selection maps obtained by CaGNet(pi). For better visualization, we use three different colors to represent the most suitable scale for each pixel.
Small scale, middle scale, and large scale are respectively denoted by blue, green, and red.
It can be observed from Fig.~\ref{viscon} that pixels in discriminative local regions (\emph{e.g.}, animal faces, small objects on the table) tend to be affected by small scale, because small-scale contextual information is sufficient for these pixels to reconstruct features. Other pixels prefer middle or large scale since they require contextual information within larger scale. By discovering the most suitable scale of context for each pixel, the context selector plays a critical role in our contextual module. More analyses of context selector and contextual model $CM$ can be found in \cite{CaGNet}.

\noindent\textbf{Category patch generation:} 
To qualitatively verify the effectiveness of category patch generation using PixelCNN, we visualize some $3\times 3$ mixed patches (constrained PixelCNN) and mixed patches (constrained random). As mentioned in Sec.~\ref{sec:blocka}, the constraints mean that we preset $\lambda_f = 5$ unseen pixels in each patch and filter out the patches containing more than $3$ categories.
For each preset unseen category (the first $\lambda_f$ pixels in the patch are set as this unseen category), we first generate $100$ mixed patches (constrained random) and $100$ mixed patches (constrained PixelCNN). Then, we calculate the logarithmic likelihood of these mixed patches according to (\ref{pc}), which reflects the probability of authenticity and reasonableness of each category patch. Finally, for each preset unseen category, we show the category patch with the largest logarithmic likelihood from mixed patches (constrained random) and mixed patches (constrained PixelCNN) respectively in Fig.~\ref{reasonable}.
We find that our category patch generation method (``constrained PixelCNN") beats random method (``constrained random") in spatial object arrangement and object-level continuities.
For spatial object arrangement, in the fifth column of Fig.~\ref{reasonable}, ``tv monitor'', ``sofa'', and ``cat'' are adjacent in the patch generated by our method, which is reasonable and common in reality. However, ``tv monitor'', ``cow'', and ``bird'' are adjacent when using random method, which rarely happens in the real world. For object-level continuities, in the first column of ``constrained random", two ``cat'' pixels are split by ``train''. On the contrary, the pixels of the same category are usually gathered together in our method, which demonstrates better object-level continuities.

\section{Conclusion}

In our proposed CaGNet for zero-shot semantic segmentation, we utilize contextual information to generate diverse and context-aware features by unifying segmentation and feature generation.
We have successfully extended the pixel-wise feature generation and finetuning to patch-wise ones.
The effectiveness of our method has been proved via plenty of qualitative and quantitative experiments.
In the future, we will explore how to generate large feature patches with the consideration of reasonable shapes or poses of unseen objects, which may further enhance the performance of zero-shot semantic segmentation.

\bibliographystyle{IEEEtran}
\bibliography{TNNLS-2013-P-0123}

\begin{thebibliography}{10}
\providecommand{\url}[1]{#1}
\csname url@samestyle\endcsname
\providecommand{\newblock}{\relax}
\providecommand{\bibinfo}[2]{#2}
\providecommand{\BIBentrySTDinterwordspacing}{\spaceskip=0pt\relax}
\providecommand{\BIBentryALTinterwordstretchfactor}{4}
\providecommand{\BIBentryALTinterwordspacing}{\spaceskip=\fontdimen2\font plus
\BIBentryALTinterwordstretchfactor\fontdimen3\font minus
  \fontdimen4\font\relax}
\providecommand{\BIBforeignlanguage}[2]{{%
\expandafter\ifx\csname l@#1\endcsname\relax
\typeout{** WARNING: IEEEtran.bst: No hyphenation pattern has been}%
\typeout{** loaded for the language `#1'. Using the pattern for}%
\typeout{** the default language instead.}%
\else
\language=\csname l@#1\endcsname
\fi
#2}}
\providecommand{\BIBdecl}{\relax}
\BIBdecl

\bibitem{lampert2013attribute}
C.~H. Lampert, H.~Nickisch, and S.~Harmeling, ``Attribute-based classification
  for zero-shot visual object categorization,'' \emph{IEEE Transactions on
  Pattern Analysis and Machine Intelligence}, vol.~36, no.~3, pp. 453--465,
  2013.

\bibitem{10.1145/2964284.2964319}
Y.~Yang, Y.~Luo, W.~Chen, F.~Shen, J.~Shao, and H.~T. Shen, ``Zero-shot hashing
  via transferring supervised knowledge,'' in \emph{ACM Multimedia}, Amsterdam,
  Netherlands, Oct. 2016.

\bibitem{10.1145/2578726.2578746}
A.~Habibian, T.~Mensink, and C.~G.~M. Snoek, ``Composite concept discovery for
  zero-shot video event detection,'' in \emph{ACM Multimedia}, Orlando, FL,
  Oct. 2014.

\bibitem{10.1145/3240508.3240715}
T.~Long, X.~Xu, Y.~Li, F.~Shen, J.~Song, and H.~T. Shen, ``Pseudo transfer with
  marginalized corrupted attribute for zero-shot learning,'' in \emph{ACM
  Multimedia}, Seoul, Republic of Korea, Oct. 2018.

\bibitem{Li2019Leveraging}
J.~Li, M.~Jin, K.~Lu, Z.~Ding, L.~Zhu, and Z.~Huang, ``Leveraging the invariant
  side of generative zero-shot learning,'' in \emph{Proceedings of the IEEE
  Conference on Computer Vision and Pattern Recognition}, Long Beach, CA, Jun.
  2019.

\bibitem{Mandal2019Out}
D.~Mandal, S.~Narayan, S.~Dwivedi, V.~Gupta, S.~Ahmed, F.~S. Khan, and L.~Shao,
  ``Out-of-distribution detection for generalized zero-shot action
  recognition,'' in \emph{Proceedings of the IEEE Conference on Computer Vision
  and Pattern Recognition}, Long Beach, CA, Jun. 2019.

\bibitem{zhang2018triple}
H.~Zhang, Y.~Long, Y.~Guan, and L.~Shao, ``Triple verification network for
  generalized zero-shot learning,'' \emph{IEEE Transactions on Image
  Processing}, vol.~28, no.~1, pp. 506--517, 2018.

\bibitem{guo2017zero}
Y.~Guo, G.~Ding, J.~Han, and Y.~Gao, ``Zero-shot learning with transferred
  samples,'' \emph{IEEE Transactions on Image Processing}, vol.~26, no.~7, pp.
  3277--3290, 2017.

\bibitem{rahman2018unified}
S.~Rahman, S.~Khan, and F.~Porikli, ``A unified approach for conventional
  zero-shot, generalized zero-shot, and few-shot learning,'' \emph{IEEE
  Transactions on Image Processing}, vol.~27, no.~11, pp. 5652--5667, 2018.

\bibitem{huynh2020fine}
D.~Huynh and E.~Elhamifar, ``Fine-grained generalized zero-shot learning via
  dense attribute-based attention,'' in \emph{Proceedings of the IEEE
  Conference on Computer Vision and Pattern Recognition}, Jun. 2020.

\bibitem{yu2020episode}
Y.~Yu, Z.~Ji, J.~Han, and Z.~Zhang, ``Episode-based prototype generating
  network for zero-shot learning,'' in \emph{Proceedings of the IEEE Conference
  on Computer Vision and Pattern Recognition}, Jun. 2020.

\bibitem{7883945}
Y.~{Cheng}, X.~{Qiao}, X.~{Wang}, and Q.~{Yu}, ``Random forest classifier for
  zero-shot learning based on relative attribute,'' \emph{IEEE Transactions on
  Neural Networks and Learning Systems}, vol.~29, no.~5, pp. 1662--1674, 2018.

\bibitem{long2015fully}
J.~Long, E.~Shelhamer, and T.~Darrell, ``Fully convolutional networks for
  semantic segmentation,'' in \emph{Proceedings of the IEEE Conference on
  Computer Vision and Pattern Recognition}, Boston, MA, Jun. 2015.

\bibitem{zhao2017pyramid}
H.~Zhao, J.~Shi, X.~Qi, X.~Wang, and J.~Jia, ``Pyramid scene parsing network,''
  in \emph{Proceedings of the IEEE Conference on Computer Vision and Pattern
  Recognition}, Honolulu, HI, Jul. 2017.

\bibitem{chen2018deeplab}
L.-C. Chen, G.~Papandreou, I.~Kokkinos, K.~Murphy, and A.~L. Yuille, ``Deeplab:
  Semantic image segmentation with deep convolutional nets, atrous convolution,
  and fully connected crfs,'' \emph{IEEE Transactions on Pattern Analysis and
  Machine Intelligence}, vol.~40, no.~4, pp. 834--848, 2018.

\bibitem{ronneberger2015u}
O.~Ronneberger, P.~Fischer, and T.~Brox, ``U-net: Convolutional networks for
  biomedical image segmentation,'' in \emph{International Conference on Medical
  Image Computing and Computer-assisted Intervention}, Munich, Germany, Oct.
  2015, pp. 234--241.

\bibitem{Lin2016RefineNet}
G.~Lin, A.~Milan, C.~Shen, and I.~Reid, ``Refinenet: Multi-path refinement
  networks for high-resolution semantic segmentation,'' in \emph{Proceedings of
  the IEEE Conference on Computer Vision and Pattern Recognition}, Honolulu,
  HI, Jul. 2017.

\bibitem{GaoLearning}
Z.~Zhang, A.~Chen, L.~Xie, J.~Yu, and S.~Gao, ``Learning semantics-aware
  distance map with semantics layering network for amodal instance
  segmentation,'' in \emph{ACM Multimedia}, Nice, France, Oct. 2019.

\bibitem{fu2019stacked}
J.~Fu, J.~Liu, Y.~Wang, J.~Zhou, C.~Wang, and H.~Lu, ``Stacked deconvolutional
  network for semantic segmentation,'' \emph{IEEE Transactions on Image
  Processing}, pp. 1--1, 2019.

\bibitem{bi2020polarimetric}
H.~Bi, L.~Xu, X.~Cao, Y.~Xue, and Z.~Xu, ``Polarimetric sar image semantic
  segmentation with 3d discrete wavelet transform and markov random field,''
  \emph{IEEE Transactions on Image Processing}, vol.~29, pp. 6601--6614, 2020.

\bibitem{8320527}
J.~{Liu}, Y.~{Wang}, Y.~{Li}, J.~{Fu}, J.~{Li}, and H.~{Lu}, ``Collaborative
  deconvolutional neural networks for joint depth estimation and semantic
  segmentation,'' \emph{IEEE Transactions on Neural Networks and Learning
  Systems}, vol.~29, no.~11, pp. 5655--5666, 2018.

\bibitem{9154612}
J.~{Fu}, J.~{Liu}, J.~{Jiang}, Y.~{Li}, Y.~{Bao}, and H.~{Lu}, ``Scene
  segmentation with dual relation-aware attention network,'' \emph{IEEE
  Transactions on Neural Networks and Learning Systems}, pp. 1--14, 2020.

\bibitem{bucher2019zero}
M.~Bucher, T.-H. Vu, M.~Cord, and P.~P{\'e}rez, ``Zero-shot semantic
  segmentation,'' in \emph{Advances in Neural Information Processing Systems},
  Vancouver, Canada, May 2019.

\bibitem{mikolov2013distributed}
T.~Mikolov, I.~Sutskever, K.~Chen, G.~S. Corrado, and J.~Dean, ``Distributed
  representations of words and phrases and their compositionality,'' in
  \emph{Advances in Neural Information Processing Systems}, Lake Tahoe, Nevada,
  United States, Dec. 2013.

\bibitem{kato2019zero}
N.~Kato, T.~Yamasaki, and K.~Aizawa, ``Zero-shot semantic segmentation via
  variational mapping,'' in \emph{Proceedings of the IEEE International
  Conference on Computer Vision Workshops}, Seoul, Korea, Oct. 2019.

\bibitem{xian2019semantic}
Y.~Xian, S.~Choudhury, Y.~He, B.~Schiele, and Z.~Akata, ``Semantic projection
  network for zero-and few-label semantic segmentation,'' in \emph{Proceedings
  of the IEEE Conference on Computer Vision and Pattern Recognition}, Long
  Beach, CA, Jun. 2019.

\bibitem{liu2020learning}
H.~Liu, Y.~Wang, J.~Zhao, G.~Yang, and F.~Lv, ``Learning unbiased zero-shot
  semantic segmentation networks via transductive transfer,'' \emph{arXiv
  preprint arXiv:2007.00515}, 2020.

\bibitem{2018Attribute}
S.~Yang, Y.~Shi, Y.~Wang, J.~Wang, and Z.~Fei, ``Attribute driven zero-shot
  classification and segmentation,'' in \emph{2018 IEEE International
  Conference on Multimedia Expo Workshops}, San Diego, USA, Jul. 2018.

\bibitem{lv2020learning}
F.~Lv, H.~Liu, Y.~Wang, J.~Zhao, and G.~Yang, ``Learning unbiased zero-shot
  semantic segmentation networks via transductive transfer,'' \emph{IEEE Signal
  Processing Letters}, vol.~27, pp. 1640--1644, 2020.

\bibitem{li2020consistent}
P.~Li, Y.~Wei, and Y.~Yang, ``Consistent structural relation learning for
  zero-shot segmentation,'' in \emph{Advances in Neural Information Processing
  Systems}, Virtual-only, Dec. 2020.

\bibitem{hu2020uncertainty}
P.~Hu, S.~Sclaroff, and K.~Saenko, ``Uncertainty-aware learning for zero-shot
  semantic segmentation,'' in \emph{Advances in Neural Information Processing
  Systems}, Virtual-only, Dec. 2020.

\bibitem{isola2017image}
P.~Isola, J.-Y. Zhu, T.~Zhou, and A.~A. Efros, ``Image-to-image translation
  with conditional adversarial networks,'' in \emph{Proceedings of the IEEE
  Conference on Computer Vision and Pattern Recognition}, Honolulu, HI, Jul.
  2017.

\bibitem{mathieu2015deep}
M.~Mathieu, C.~Couprie, and Y.~LeCun, ``Deep multi-scale video prediction
  beyond mean square error,'' in \emph{International Conference on Learning
  Representations}, San Juan, Puerto Rico, May 2016.

\bibitem{pathak2016context}
D.~Pathak, P.~Krahenbuhl, J.~Donahue, T.~Darrell, and A.~A. Efros, ``Context
  encoders: Feature learning by inpainting,'' in \emph{Proceedings of the IEEE
  conference on computer vision and pattern recognition}, Las Vegas, NV, Jun.
  2016, pp. 2536--2544.

\bibitem{NeurIPS2017_6650}
J.-Y. Zhu, R.~Zhang, D.~Pathak, T.~Darrell, A.~A. Efros, O.~Wang, and
  E.~Shechtman, ``Toward multimodal image-to-image translation,'' in
  \emph{Advances in Neural Information Processing Systems}, Long Beach, CA,
  Dec. 2017.

\bibitem{mottaghi2014role}
R.~Mottaghi, X.~Chen, X.~Liu, N.-G. Cho, S.-W. Lee, S.~Fidler, R.~Urtasun, and
  A.~Yuille, ``The role of context for object detection and semantic
  segmentation in the wild,'' in \emph{Proceedings of the IEEE Conference on
  Computer Vision and Pattern Recognition}, Columbus, OH, Jun. 2014.

\bibitem{pixelrnn}
A.~v.~d. Oord, N.~Kalchbrenner, and K.~Kavukcuoglu, ``Pixel recurrent neural
  networks,'' in \emph{Proceedings of the IEEE International Conference on
  Machine Learning}, New York, USA, Jun. 2016.

\bibitem{CaGNet}
Z.~Gu, S.~Zhou, L.~Niu, Z.~Zhao, and L.~Zhang, ``Context-aware feature
  generation for zero-shot semantic segmentation,'' in \emph{ACM Multimedia},
  Seattle, United States, Oct. 2020.

\bibitem{oh2017exploiting}
S.~J. Oh, R.~Benenson, A.~Khoreva, Z.~Akata, M.~Fritz, and B.~Schiele,
  ``Exploiting saliency for object segmentation from image level labels,'' in
  \emph{Proceedings of the IEEE Conference on Computer Vision and Pattern
  Recognition}, Honolulu, HI, Jul. 2017.

\bibitem{papandreou2015weakly}
G.~Papandreou, L.-C. Chen, K.~P. Murphy, and A.~L. Yuille, ``Weakly-and
  semi-supervised learning of a deep convolutional network for semantic image
  segmentation,'' in \emph{Proceedings of the IEEE International Conference on
  Computer Vision}, Santiago, Chile, Dec. 2015.

\bibitem{Yao2015Semantic}
X.~Yao, J.~Han, C.~Gong, and G.~Lei, ``Semantic segmentation based on stacked
  discriminative autoencoders and context-constrained weakly supervised
  learning,'' in \emph{ACM Multimedia}, Sydney, Australia, Oct. 2015.

\bibitem{khoreva2017simple}
A.~Khoreva, R.~Benenson, J.~Hosang, M.~Hein, and B.~Schiele, ``Simple does it:
  Weakly supervised instance and semantic segmentation,'' in \emph{Proceedings
  of the IEEE Conference on Computer Vision and Pattern Recognition}, Honolulu,
  HI, Jul. 2017.

\bibitem{GraphNet}
M.~Pu, Y.~Huang, Q.~Guan, and Q.~Zou, ``Graphnet: Learning image pseudo
  annotations for weakly-supervised semantic segmentation,'' in \emph{ACM
  Multimedia}, Seoul, Republic of Korea, Oct. 2018.

\bibitem{lin2016scribblesup}
D.~Lin, J.~Dai, J.~Jia, K.~He, and J.~Sun, ``Scribblesup: Scribble-supervised
  convolutional networks for semantic segmentation,'' in \emph{Proceedings of
  the IEEE Conference on Computer Vision and Pattern Recognition}, Las Vegas,
  NV, Jul. 2016.

\bibitem{Lampert2009Learning}
C.~H. Lampert, H.~Nickisch, and S.~Harmeling, ``Learning to detect unseen
  object classes by between-class attribute transfer,'' in \emph{Proceedings of
  the IEEE Conference on Computer Vision and Pattern Recognition}, Miami, FL,
  Jun. 2009.

\bibitem{akata2015label}
Z.~Akata, F.~Perronnin, Z.~Harchaoui, and C.~Schmid, ``Label-embedding for
  image classification,'' \emph{IEEE Transactions on Pattern Analysis and
  Machine Intelligence}, vol.~38, no.~7, pp. 1425--1438, 2015.

\bibitem{frome2013devise}
A.~Frome, G.~S. Corrado, J.~Shlens, S.~Bengio, J.~Dean, M.~Ranzato, and
  T.~Mikolov, ``Devise: A deep visual-semantic embedding model,'' in
  \emph{Advances in Neural Information Processing Systems}, Lake Tahoe, Nevada,
  United States, Dec. 2013.

\bibitem{romera2015embarrassingly}
B.~Romera-Paredes and P.~Torr, ``An embarrassingly simple approach to zero-shot
  learning,'' in \emph{Proceedings of the IEEE International Conference on
  Machine Learning}, Lille, France, Jul. 2015.

\bibitem{fu2015transductive}
Y.~Fu, T.~M. Hospedales, T.~Xiang, and S.~Gong, ``Transductive multi-view
  zero-shot learning,'' \emph{IEEE Transactions on Pattern Analysis and Machine
  Intelligence}, vol.~37, no.~11, pp. 2332--2345, 2015.

\bibitem{liu2020iterative}
M.~Liu, L.~Qu, L.~Nie, M.~Liu, L.~Duan, and B.~Chen, ``Iterative local-global
  collaboration learning towards one-shot video person re-identification,''
  \emph{IEEE Transactions on Image Processing}, vol.~29, pp. 9360--9372, 2020.

\bibitem{shaban2017one}
A.~Shaban, S.~Bansal, Z.~Liu, I.~Essa, and B.~Boots, ``One-shot learning for
  semantic segmentation,'' \emph{arXiv preprint arXiv:1709.03410}, 2017.

\bibitem{zhang2020sg}
X.~Zhang, Y.~Wei, Y.~Yang, and T.~S. Huang, ``Sg-one: Similarity guidance
  network for one-shot semantic segmentation,'' \emph{IEEE Transactions on
  Cybernetics}, vol.~50, no.~9, pp. 3855--3865, 2020.

\bibitem{zhang2019pyramid}
C.~Zhang, G.~Lin, F.~Liu, J.~Guo, Q.~Wu, and R.~Yao, ``Pyramid graph networks
  with connection attentions for region-based one-shot semantic segmentation,''
  in \emph{Proceedings of the IEEE International Conference on Computer Vision
  Workshops}, Seoul, Korea, Oct. 2019.

\bibitem{caelles2017one}
S.~Caelles, K.-K. Maninis, J.~Pont-Tuset, L.~Leal-Taix{\'e}, D.~Cremers, and
  L.~Van~Gool, ``One-shot video object segmentation,'' in \emph{Proceedings of
  the IEEE conference on computer vision and pattern recognition}, Hawaii,USA,
  Jul. 2017, pp. 221--230.

\bibitem{xian2018feature}
Y.~Xian, T.~Lorenz, B.~Schiele, and Z.~Akata, ``Feature generating networks for
  zero-shot learning,'' in \emph{Proceedings of the IEEE Conference on Computer
  Vision and Pattern Recognition}, Salt Lake City, UT, Jun. 2018.

\bibitem{FelixMulti}
R.~Felix, B.~G.~V. Kumar, I.~Reid, and G.~Carneiro, ``Multi-modal
  cycle-consistent generalized zero-shot learning,'' in \emph{Proceedings of
  the European Conference on Computer Vision}, Munich,Germany, Sep. 2018.

\bibitem{xian2019f}
Y.~Xian, S.~Sharma, B.~Schiele, and Z.~Akata, ``f-vaegan-d2: A feature
  generating framework for any-shot learning,'' in \emph{Proceedings of the
  IEEE Conference on Computer Vision and Pattern Recognition}, Long Beach, CA,
  Jun. 2019.

\bibitem{Mert2019Gradient}
M.~B. Sariyildiz and R.~G. Cinbis, ``Gradient matching generative networks for
  zero-shot learning,'' in \emph{Proceedings of the IEEE Conference on Computer
  Vision and Pattern Recognition}, Long Beach, CA, Jun. 2019.

\bibitem{goodfellow2014generative}
I.~Goodfellow, J.~Pouget-Abadie, M.~Mirza, B.~Xu, D.~Warde-Farley, S.~Ozair,
  A.~Courville, and Y.~Bengio, ``Generative adversarial nets,'' in
  \emph{Advances in Neural Information Processing Systems}, Montreal, Canada,
  Dec. 2014.

\bibitem{brock2018large}
A.~Brock, J.~Donahue, and K.~Simonyan, ``Large scale gan training for high
  fidelity natural image synthesis,'' \emph{arXiv preprint arXiv:1809.11096},
  2018.

\bibitem{ledig2017photo}
C.~Ledig, L.~Theis, F.~Husz{\'a}r, J.~Caballero, A.~Cunningham, A.~Acosta,
  A.~Aitken, A.~Tejani, J.~Totz, Z.~Wang \emph{et~al.}, ``Photo-realistic
  single image super-resolution using a generative adversarial network,'' in
  \emph{Proceedings of the IEEE conference on computer vision and pattern
  recognition}, Hawaii,USA, Jul. 2017, pp. 4681--4690.

\bibitem{MaoLeast}
X.~Mao, Q.~Li, H.~Xie, R.~Y.~K. Lau, Z.~Wang, and S.~P. Smolley, ``Least
  squares generative adversarial networks,'' in \emph{Proceedings of the IEEE
  International Conference on Computer Vision}, Venice, Italy, Oct. 2017.

\bibitem{zhang2017stackgan}
H.~Zhang, T.~Xu, H.~Li, S.~Zhang, X.~Wang, X.~Huang, and D.~N. Metaxas,
  ``Stackgan: Text to photo-realistic image synthesis with stacked generative
  adversarial networks,'' in \emph{Proceedings of the IEEE international
  conference on computer vision}, Venice, Italy, Oct. 2017, pp. 5907--5915.

\bibitem{zhang2019gain}
J.~Zhang, L.~Niu, D.~Yang, L.~Kang, Y.~Li, W.~Zhao, and L.~Zhang, ``Gain:
  Gradient augmented inpainting network for irregular holes,'' in
  \emph{Proceedings of the 27th ACM International Conference on Multimedia},
  Nice, France, Oct. 2019, pp. 1870--1878.

\bibitem{kipf2016semi-supervised}
T.~Kipf and M.~Welling, ``Semi-supervised classification with graph
  convolutional networks,'' in \emph{International Conference on Learning
  Representations}, Toulon, France, Apr. 2017.

\bibitem{xu2018pad-net:}
D.~Xu, W.~Ouyang, X.~Wang, and N.~Sebe, ``Pad-net: Multi-tasks guided
  prediction-and-distillation network for simultaneous depth estimation and
  scene parsing,'' in \emph{Proceedings of the European Conference on Computer
  Vision}, Munich,Germany, Sep. 2018, pp. 675--684.

\bibitem{Yu2015Multi}
F.~Yu and V.~Koltun, ``Multi-scale context aggregation by dilated
  convolutions,'' in \emph{International Conference on Learning
  Representations}, San Juan, Puerto Rico, May 2016.

\bibitem{hu2018gather}
J.~Hu, L.~Shen, S.~Albanie, G.~Sun, and A.~Vedaldi, ``Gather-excite: Exploiting
  feature context in convolutional neural networks,'' in \emph{Advances in
  Neural Information Processing Systems}, Montral,Canada, Dec. 2018.

\bibitem{kingma2013auto}
D.~P. Kingma and M.~Welling, ``Auto-encoding variational bayes,'' \emph{arXiv
  preprint arXiv:1312.6114}, 2013.

\bibitem{resnet-101}
K.~He, X.~Zhang, S.~Ren, and J.~Sun, ``Deep residual learning for image
  recognition,'' in \emph{Proceedings of the IEEE Conference on Computer Vision
  and Pattern Recognition}, Las Vegas, NV, Jun. 2016.

\bibitem{ILSVRC15}
O.~Russakovsky, J.~Deng, H.~Su, J.~Krause, S.~Satheesh, S.~Ma, Z.~Huang,
  A.~Karpathy, A.~Khosla, M.~Bernstein, A.~C. Berg, and L.~Fei-Fei, ``Imagenet
  large scale visual recognition challenge,'' \emph{IJCV}, vol. 115, no.~3, pp.
  211--252, 2015.

\bibitem{EveringhamThePascalVisual}
M.~Everingham, S.~M.~A. Eslami, L.~Van~Gool, C.~K.~I. Williams, J.~Winn, and
  A.~Zisserman, ``The pascal visual object classes challenge: A
  retrospective,'' \emph{International Journal of Computer Vision}, vol. 111,
  no.~1, pp. 98--136, 2015.

\bibitem{caesar2018coco}
H.~Caesar, J.~Uijlings, and V.~Ferrari, ``Coco-stuff: Thing and stuff classes
  in context,'' in \emph{Proceedings of the IEEE Conference on Computer Vision
  and Pattern Recognition}, Salt Lake City, UT, Jun. 2018.

\bibitem{Hariharan2011Semantic}
B.~Hariharan, P.~Arbelaez, L.~D. Bourdev, S.~Maji, and J.~Malik, ``Semantic
  contours from inverse detectors,'' in \emph{Proceedings of the IEEE
  International Conference on Computer Vision}, Barcelona, Spain, Nov. 2011.

\bibitem{joulin2017bag}
A.~Joulin, E.~Grave, P.~Bojanowski, and T.~Mikolov, ``Bag of tricks for
  efficient text classification,'' in \emph{Conference of the European Chapter
  of the Association for Computational Linguistics}, Valencia, Spain, Apr.
  2017.

\bibitem{luo2016understanding}
W.~Luo, Y.~Li, R.~Urtasun, and R.~S. Zemel, ``Understanding the effective
  receptive field in deep convolutional neural networks,'' in \emph{Advances in
  Neural Information Processing Systems}, Barcelona, Spain, Dec. 2016.

\end{thebibliography}

\begin{IEEEbiography}[{\includegraphics[width=1in,height=1.25in,clip,keepaspectratio]{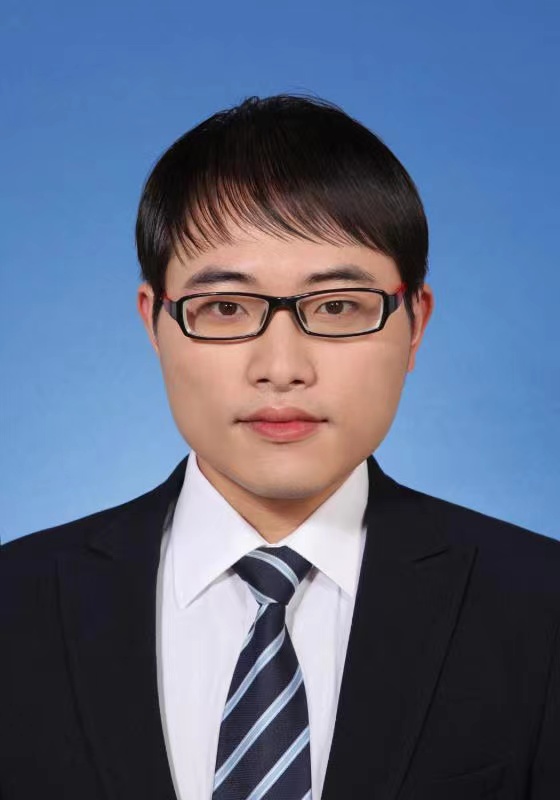}}]{Zhanxuan Gu}
Zhangxuan Gu is currently a PHD student in Computer Science and Engineering Department at Shanghai Jiao Tong University, Shanghai, China. Before that, he obtained his B.E. degree on Mathematics from the Shanghai Jiao Tong University. His current research interests include machine learning, deep learning, and computer vision.
\end{IEEEbiography}

\begin{IEEEbiography}[{\includegraphics[width=1in,height=1.25in,clip,keepaspectratio]{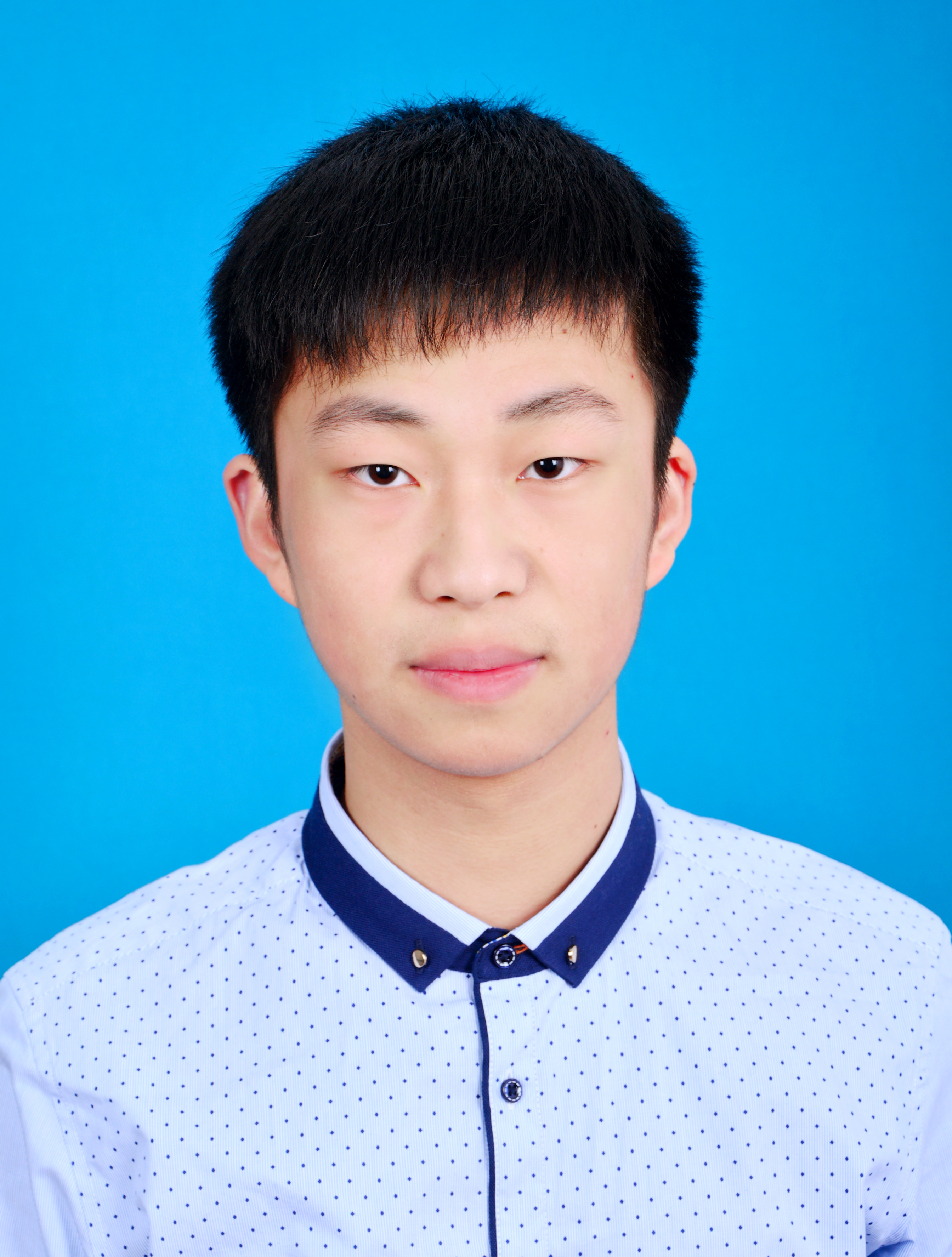}}]{Siyuan Zhou}
Siyuan Zhou is currently studying for a Master's degree in Computer Science and Engineering Department at Shanghai Jiao Tong University, Shanghai, China. Prior to that, he obtained his B.E. degree from Shanghai Jiao Tong University, Shanghai, China in 2020. His current research interests include machine learning, deep learning, and computer vision.
\end{IEEEbiography}

\begin{IEEEbiography}[{\includegraphics[width=1in,height=1.25in,clip,keepaspectratio]{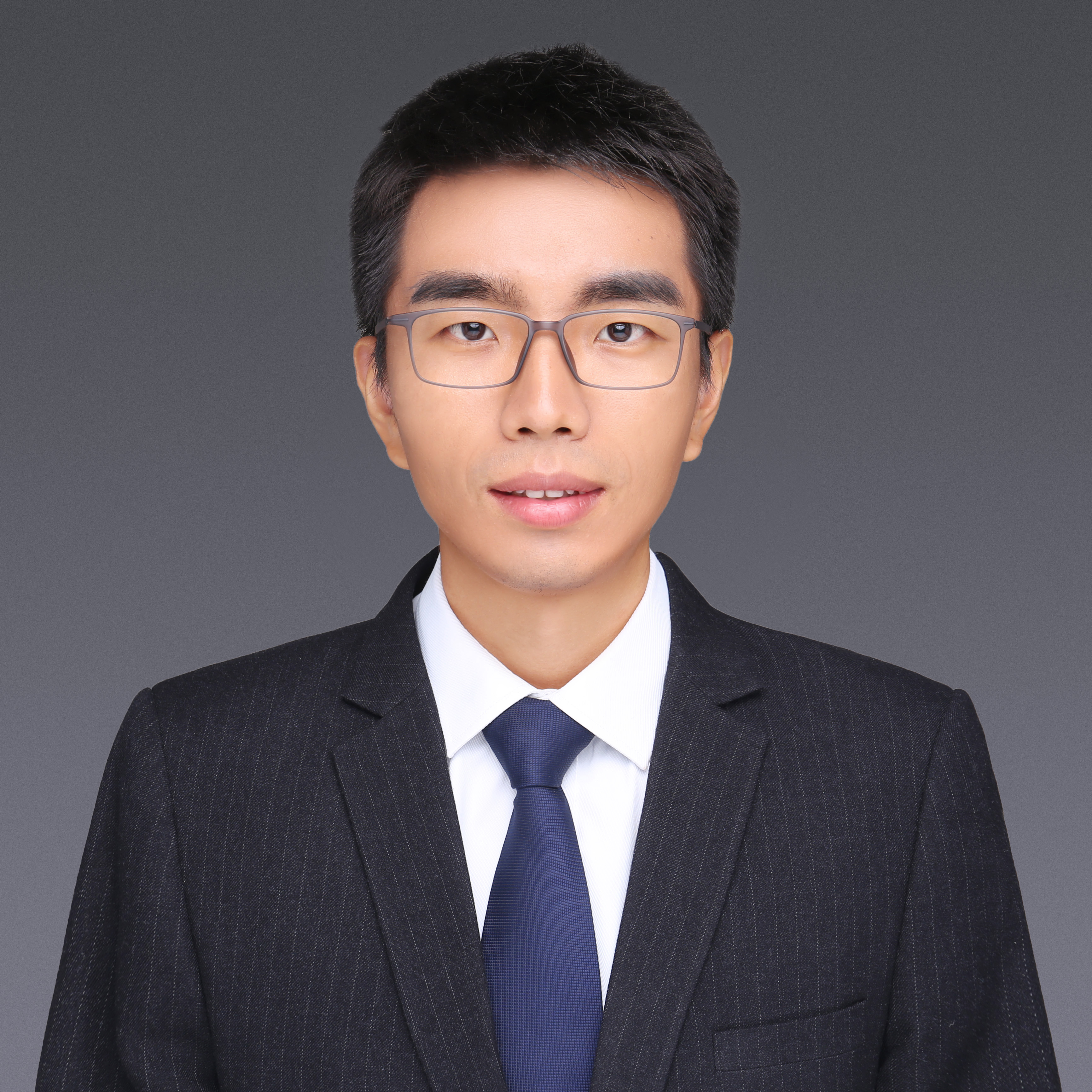}}]{Li Niu}
Li Niu is currently an associate professor in Computer Science and Engineering Department at Shanghai Jiao Tong University, Shanghai, China. Before joining Shanghai Jiao Tong University, he was a postdoctoral associate at Rice University in Houston, TX, USA. Prior to that, he obtained his B.E. degree from the University of Science and Technology of China, Hefei, China in 2011 and Ph.D. degree from Nanyang Technological University, Singapore in 2017. His current research interests include machine learning, deep learning, and computer vision.
\end{IEEEbiography}

\begin{IEEEbiography}[{\includegraphics[width=1in,height=1.25in,clip,keepaspectratio]{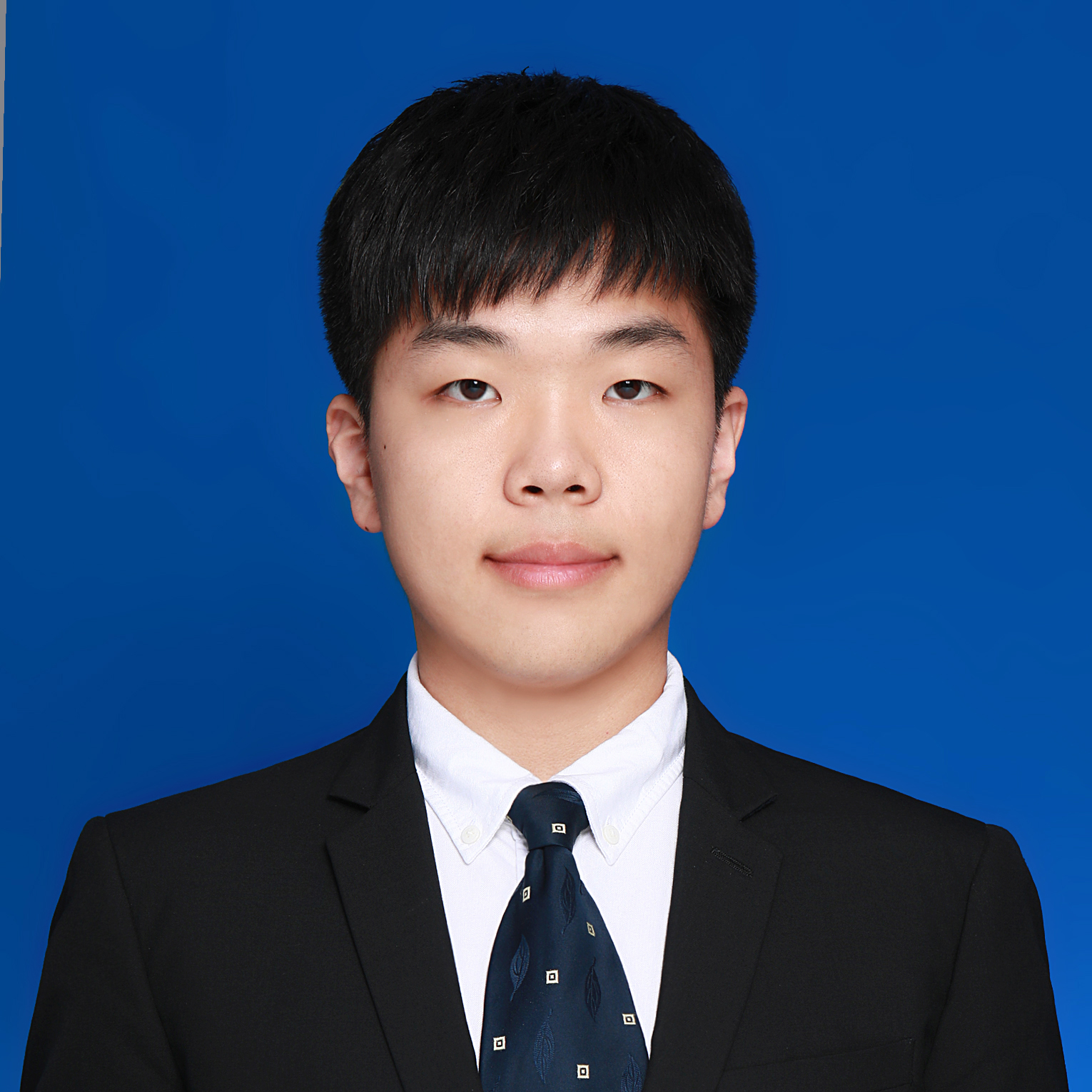}}]{Zihan Zhao}
Zihan Zhao is currently a Ph.D student in Computer Science and Engineering Department at Shanghai Jiao Tong University, Shanghai, China. Before this, he obtained his B.S. degree of chemistry from Shanghai Jiao Tong University in 2020. His current research interests include machine learning, deep learning, and natural language processing.
\end{IEEEbiography}

\begin{IEEEbiography}[{\includegraphics[width=1in,height=1.25in,clip,keepaspectratio]{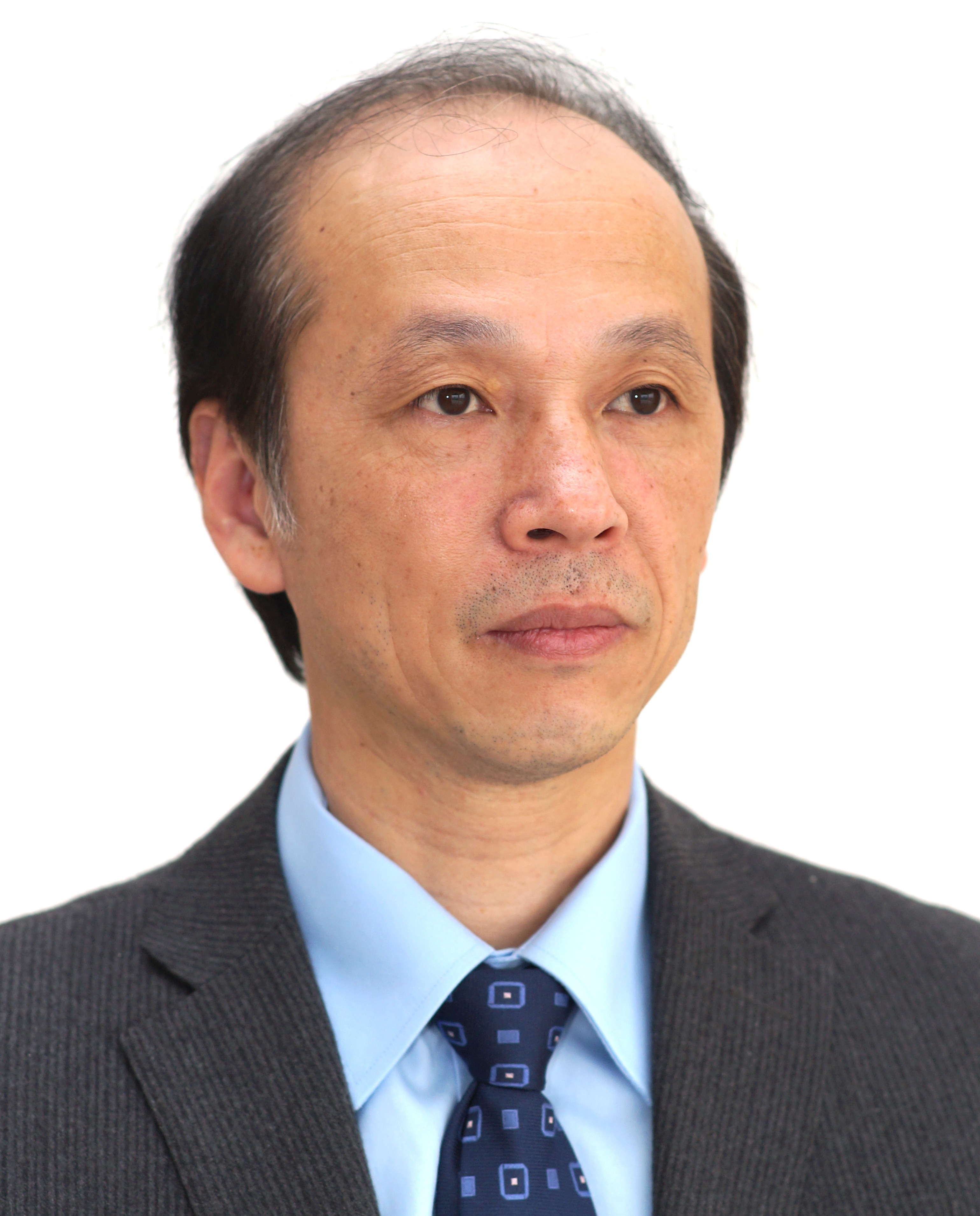}}]{Liqing Zhang}
Liqing Zhang received the Ph.D. degree from Zhongshan University, Guangzhou, China, in 1988. He was promoted to full professor position in 1995 at South China University of Technology. He worked as a research scientist in RIKEN Brain Science Institute, Japan from 1997 to 2002. Since September, 2002, he has been a Professor with Department of Computer Science and Engineering, Shanghai Jiao Tong University, Shanghai, China. His current research interests cover computational theory for cortical networks, visual cognitive representation and inference, statistical learning. He has published more than 250 papers in journals and international conferences.
\end{IEEEbiography}

\appendices

\end{document}